\documentclass[twoside,11pt]{article}
\usepackage{jair, theapa, rawfonts}
\usepackage{xcolor}
\usepackage{amsmath}
\usepackage{amssymb}
\usepackage{enumitem}
\usepackage{graphicx}
\usepackage{comment}
\usepackage{booktabs}
\usepackage{mdframed}
\usepackage{amsthm}
\usepackage{subcaption}
\usepackage{xspace}

\usepackage[normalem]{ulem}
\usepackage{xcolor}

\newtheorem{definition}{Definition}
\newtheorem{proposition}{Proposition}
\newtheorem{remark}{Remark}
\newtheorem{example}{Example} 

\usepackage{tikz}
\usetikzlibrary{calc, automata, arrows.meta, positioning, shadows, fit, shapes.misc, decorations.pathreplacing, backgrounds, math, pgfplots.groupplots, shapes.geometric}

\newmdenv[
  backgroundcolor=TACcyanwashed,
  shadow=false,
  hidealllines=true,
  leftline=true,
  linecolor=TACcyan,
  linewidth=3pt,
  skipabove=4pt,
  skipbelow=4pt,
  innerleftmargin=10pt,
  innerrightmargin=10pt,
  innertopmargin=0pt,
  innerbottommargin=8pt
]{bluebox}

\newmdenv[
  backgroundcolor=grayfilling,
  shadow=false,
  hidealllines=true,
  leftline=true,
  linecolor=gray,
  linewidth=3pt,
  skipabove=4pt,
  skipbelow=4pt,
  innerleftmargin=10pt,
  innerrightmargin=10pt,
  innertopmargin=0pt,
  innerbottommargin=8pt
]{graybox}

\newmdenv[
  backgroundcolor=cherryred!10,
  shadow=false,
  hidealllines=true,
  leftline=true,
  linecolor=cherryred,
  linewidth=3pt,
  skipabove=4pt,
  skipbelow=4pt,
  innerleftmargin=10pt,
  innerrightmargin=10pt,
  innertopmargin=0pt,
  innerbottommargin=8pt
]{redbox}

\newmdtheoremenv[
  backgroundcolor=grayfilling,
  hidealllines=true,
  leftline=true,
  linecolor=gray,
  linewidth=3pt,
  skipabove=6pt,
  skipbelow=6pt,
  innerleftmargin=10pt,
  innerrightmargin=10pt,
  innertopmargin=2pt,
  innerbottommargin=10pt
]{lemma}{Lemma}

\newmdtheoremenv[
  backgroundcolor=grayfilling,
  hidealllines=true,
  leftline=true,
  linecolor=gray,
  linewidth=3pt,
  skipabove=6pt,
  skipbelow=6pt,
  innerleftmargin=10pt,
  innerrightmargin=10pt,
  innertopmargin=2pt,
  innerbottommargin=10pt
]{problem}{Problem}

\newmdtheoremenv[
  backgroundcolor=cherryred!10,
  hidealllines=true,
  leftline=true,
  linecolor=cherryred,
  linewidth=3pt,
  skipabove=6pt,
  skipbelow=6pt,
  innerleftmargin=10pt,
  innerrightmargin=10pt,
  innertopmargin=2pt,
  innerbottommargin=10pt
]{assumption}{Assumption}

\newmdtheoremenv[
  backgroundcolor=TACcyanwashed,
  hidealllines=true,
  leftline=true,
  linecolor=TACcyan,
  linewidth=3pt,
  skipabove=6pt,
  skipbelow=6pt,
  innerleftmargin=10pt,
  innerrightmargin=10pt,
  innertopmargin=2pt,
  innerbottommargin=10pt
]{theorem}{Theorem}

\definecolor{TACcyan}{RGB}{0,141,225}
\colorlet{TACcyanwashed}{TACcyan!10}
\definecolor{lightgreen}{rgb}{0.67, 0.88, 0.69}
\definecolor{lightpink}{rgb}{1.0, 0.72, 0.77}
\definecolor{lightpurple}{rgb}{0.96, 0.73, 1.0}
\definecolor{lightyellow}{rgb}{0.98, 0.93, 0.37}
\usepackage{pgfplots}
\definecolor{grayfilling}{gray}{0.95} 
\definecolor{grayshadow}{gray}{0.5} 
\colorlet{cherryred}{red!80!black}
\definecolor{accent}{RGB}{248, 29, 248}

\newcommand{\ambRKHS}{\mathcal{C}} 
\newcommand{\abs}[1]{\left\vert#1\right\vert} 
\newcommand{\B}{B}
\newcommand{\Barrmax}{{\hat{\B}}} 
\newcommand{\Barrmin}{{\check{\B}}} 
\newcommand{\bigO}{\mathcal{O}} 
\newcommand{\borel}[1]{\mathcal{B}(#1)}

\newcommand{\cdotx}{\,\cdot\,}

\newcommand{\diag}{\mathrm{diag}}

\newcommand{\dilation}{\vartheta} 
\newcommand{\E}{\mathbb{E}}

\newcommand{\Hilbert}{\mathcal{H}}
\newcommand{\innerH}[3]{\langle #1, #2 \rangle_{#3}} 
\newcommand{\LTLf}{LTL\textsubscript{F}\xspace}
\newcommand{\M}{\mathbf{M}}
\newcommand{\N}{\mathbb{N}}
\newcommand{\norm}[1]{\left|\left|#1\right|\right|}
\renewcommand{\P}{\mathbb{P}}
\newcommand{\R}{\mathbb{R}}
\newcommand{\U}{\mathbb{U}}
\newcommand{\w}{{{\mathbf{w}}}} 
\newcommand{\satisfies}{\vDash}
\renewcommand{\S}{\mathbb{S}\xspace} 
\newcommand{\T}{^\top}
\newcommand{\Tr}{\mathbf{t}}
\newcommand{\p}{\mathbf{p}}
\newcommand{\X}{\mathbb{X}}

\newcommand{\Y}{\mathbb{Y}}

\newcommand{\lucid}{\textsc{Lucid}\xspace}

\graphicspath{{./Figures/}} 
\usepackage[outdir=./Figures/]{epstopdf}

\usepackage{import}
\usepackage{xifthen}
\usepackage{pdfpages}
\usepackage{transparent}

\usepackage{pgfplots} 
\usepgfplotslibrary{fillbetween} 
\pgfmathdeclarefunction{gauss}{3}{%
    \pgfmathparse{1.5*1/(#3*sqrt(2*pi))*exp(-((#1-#2)^2)/(2*#3^2))}%
}
\usepackage[outline]{contour} 
\usetikzlibrary{patterns,positioning,fit}
\pgfplotsset{compat=1.12} 

\def\titlestring{Kernel-Based Learning of Safety Barriers}
\jairheading{X}{2025}{X-X}{01/25}{X/X}
\ShortHeadings{\titlestring}
{Sch\"on, Zhong, \& Soudjani}
\firstpageno{1}

\begin{document}

\title{\titlestring}

\author{\name Oliver Sch\"on \email o.schoen2@ncl.ac.uk \\
       \addr Newcastle University, School of Computing, Newcastle upon Tyne, \\NE4 5TG, United Kingdom
       \AND
       \name Zhengang Zhong \email zhengang.zhong@warwick.ac.uk \\
       \addr University of Warwick, Department of Statistics, Coventry, \\CV4 7AL, United Kingdom
       \AND
       \name Sadegh Soudjani \email sadegh@mpi-sws.org \\
       \addr Max Planck Institute for Software Systems, Kaiserslautern, 67663, Germany\\
       University of Birmingham, Birmingham, B15 2TT, United Kingdom.}


\maketitle

\begin{abstract}
The rapid integration of AI algorithms in safety-critical applications such as autonomous driving and healthcare is raising significant concerns about the ability to meet stringent safety standards.
Traditional tools for formal safety verification struggle with the black-box nature of AI-driven systems and lack the flexibility needed to scale to the complexity of real-world applications. 
In this paper, we present a data-driven approach for safety verification and synthesis of black-box systems with discrete-time stochastic dynamics.
We employ the concept of control barrier certificates, which can guarantee safety of the system, and learn the certificate directly from a set of system trajectories. We use conditional mean embeddings to embed data from the system into a reproducing kernel Hilbert space (RKHS) and construct an RKHS ambiguity set that can be inflated to robustify the result to out-of-distribution behavior.
We provide the theoretical results on how to apply the approach to general classes of temporal logic specifications beyond safety. For the data-driven computation of safety barriers, we {leverage} a finite Fourier expansion to cast a typically intractable semi-infinite optimization problem as a linear program.
The resulting spectral barrier allows us to leverage the fast Fourier transform to generate the relaxed problem efficiently, offering a scalable yet distributionally robust framework for verifying safety.
Our work moves beyond restrictive assumptions on system dynamics and uncertainty, as demonstrated on {two} case studies including a black-box system with a neural network controller.
\end{abstract}

\section{Introduction}
\label{sec:introduction}
We are living in a truly remarkable time of technological progress powered by AI. Autonomous cars, for instance, have recently been reported to have surpassed human level safety whilst driving under controlled conditions --- an achievement supported largely by extensive and costly testing~\cite{kalra2016driving}.
Yet, achieving a level of safety sufficient to earn societal trust in AI-powered safety-critical systems presents a significant challenge.
With a rapid adoption of embodied AI across various domains including smart grids and advanced medical technologies, we are facing an unprecedented demand for safe and trustworthy AI that testing cannot meet. 
In fact, in a world governed by non-determinism, exhaustively testing embodied AI against all possible edge cases is infeasible.
To address these challenges, the ability to enforce logical constraints --- such as safety and invariance --- on AI systems is emerging as a critical capability.

Formal methods for the verification and synthesis of dynamical systems are being developed to bring forth tools for certifying whether an AI system can be trusted~\shortcite{belta2017formal,Lavaei_Survey,yin2024formal}. More precisely, rich (stochastic) models of the target system's behavior are subjected to temporal logic constraints to obtain rigorous probabilistic measures of their well-behavedness. As exact system representations are rarely available, models are often derived from data. To compensate for the arising epistemic uncertainty, formal methods are equipped to deal with model ambiguity. The formal nature of the results, however, comes at the cost of onerous assumptions and a substantial computational burden. 
In fact, formal reasoning amidst unknown dynamics is only possible if the data is supplemented with information on the regularity of the system dynamics. However, this often restricts the applicability of results to specific classes of systems, such as linear, polynomial, or control-affine systems. Extending these methods to encompass more general AI-driven systems, such as those governed by neural network (NN) controllers, could act as a catalyzer for the development of safe AI systems.

Existing work on formal methods is dominated by a bipartite narrative of approaches that are either fundamentally \emph{abstraction-based} or \emph{abstraction-free}.
Whilst the former is attributed to suffer from the excruciating \emph{curse of dimensionality}, the latter is expected to be more scalable.
At a closer examination, however, this division becomes less clear.
As the arguably most popular abstraction-free formal tool, \emph{control barrier certificates}\footnote{We refer to CBCs and discrete-time control barrier functions (CBFs) \shortcite{cosner2023generative} jointly as CBCs since their definitions are equivalent.} (CBCs) allow one to reformulate the verification of safety by proving the existence of a function --- a so-called barrier --- satisfying a given set of mathematical inequalities.
The resulting global optimization problem involving the latent system dynamics is in general a robust optimization problem featuring infinitely many constraints.
Thus, many approaches rely on some form of spatial abstraction, i.e., partitioning of the system domain into discrete states and inputs, to solve the complex optimization problem (cf.~Section~\ref{sec:divide_and_conquer}).
{To address these shortcomings, we draw from a line of research that has thus far received only little attention from the formal methods community: (reproducing) kernel methods.}

Recent advancements in machine learning (ML) have underscored the significance of representing models as functions within \emph{reproducing kernel Hilbert spaces} (RKHS) \cite{scholkopf2002learning,Berlinet2004RKHSProbStat,Steinwart2008SVM}. 
An RKHS is a Hilbert space of functions where evaluation at any point is a continuous linear operation, making the RKHS formulation well-suited for analyzing and manipulating functions in high-dimensional spaces. The RKHS framework has sparked a deeper understanding of complex ML algorithms and led to the development of tools such as \emph{conditional mean embeddings} (CMEs) \shortcite{Song2009CondEmbed,Klebanov2020RigorousCME}. CMEs provide a way to embed conditional probability measures into an RKHS, enabling the computation of conditional expectations of random variables as simple inner products. This approach is particularly powerful for analyzing the expected behavior of stochastic processes, as it captures dependencies and relationships within data without requiring explicit density estimation.

In control applications, kernel-based tools such as CMEs offer several benefits over traditional data-driven optimization techniques. They allow the reformulation of nonlinear problems as linear ones within the RKHS framework, thus simplifying computations \cite{scholkopf2002learning,Berlinet2004RKHSProbStat}. Additionally, they bypass intermediate steps such as density estimation and numerical integration, facilitating the direct computation of conditional expectations on the observed data. Unlike many existing formal methods for verification and synthesis, which often rely on structural assumptions or Lipschitz continuity, CMEs can enable the approximate reformulation of the original problem using finite data based only on a standard assumption used in CME theory. 
Paired with statistical results based on concentration inequalities, probabilistic guarantees for the latent data-generating system can be obtained from small and noisy data sets~\shortcite{li2022optimal}.
This versatility and computational efficiency make CMEs particularly appealing for formal verification and synthesis.

\subsection{Contributions}
In this work, we study utilizing the CME theory for verification and control synthesis of stochastic systems without explicit model knowledge via the concept of control barrier certificates (CBCs) {\cite{prajna2007framework}}. 
By reformulating the probabilistic CBC constraints into a data-driven optimization problem, we derive distributionally robust characterizations based on an ambiguity set of candidate transition kernels. We show how these characterizations can be extended to satisfy temporal logic specifications beyond safety using an automata representation of the specification and the concept of Streett supermartingales \shortcite{abate2024stochastic}. From a computational perspective, our central innovation lies in the application of a Fourier expansion to the barrier function, yielding a computationally efficient and expressive \emph{Fourier control barrier certificate} 
We show that for safety verification, this approach collapses the complex CME term into a tractable spectral representation, enabling constraint evaluation at equidistant sample lattices via the fast Fourier transform (FFT).
The resulting spectral formulation reduces the semi-infinite program to a finitely constrained \emph{linear program} (LP), leveraging bounding results for trigonometric polynomials~\cite{pfister2018bounding}.
Our method relaxes restrictive assumptions of existing approaches and scales favorably across benchmarks.

A subset of the results of this paper was published in the conference paper by \citeA{schon2024DRObarrier}. This manuscript provides substantial extensions of previous results along the following directions.
\begin{itemize}
    \item[(a)] We generalize the theoretical framework to handle non-autonomous dynamics, yielding data-driven CBC conditions based on a general standard assumption used in CME theory.
    This removes the need for restrictive assumptions prevalent in related work (cf. related work in Section~\ref{sec:relatedWork}). 
    \item[(b)] In order to address temporal logic specifications beyond safety, we provide robust inequalities based on CME theory, an automata representation of the specification, and Streett supermartingales, by raising appropriate assumptions on the system.
    \item[(c)] With focus on safety verification, we provide a detailed study of robust optimization and semi-infinite programming techniques applicable to the derived program. These include monolithic and divide-and-conquer strategies, which we juxtapose based on their theoretical complexity and our experimental findings. 
    \item[(d)] For the general squared-exponential kernel and safety verification, we develop a truncated Fourier expansion that eliminates the reliance on large amounts of spatial support vectors, {yielding a computationally efficient method.}
    A scalable sampling-based scheme, leveraging the FFT, enables practical LP reformulations for complex stochastic systems.
    \item[(e)] The newly provided benchmarks include complex safety specifications and neural network controllers.
\end{itemize}

\subsection{Paper Organization}
The rest of the paper is organized as follows.
After a brief review of related work, we present the preliminaries and problem statement in Section~\ref{sec:preAndProblem}.
The concepts of RKHS and kernel mean embeddings for (conditional) probability measures are introduced in Section~\ref{sec:RKHS}.
We derive inequality constraints for computing CBCs directly from data in Section~\ref{sec:ddcbc}.
The application of kernel mean embeddings to general classes of temporal logic specifications beyond safety is discussed in Section~\ref{sec:LTL}.
With a focus on safety specifications, we characterize the resulting semi-infinite problem and provide a thorough comparison of possible solutions in Section~\ref{sec:probCharacterizationAndComparison}.
In Section~\ref{sec:algorithm}, we present a spectral approach based on a spectral Fourier expansion that admits a solution of the data-driven verification problem as an LP.
We demonstrate the performance of the proposed approach on {two} benchmarks followed up by concluding remarks in Sections~\ref{sec:numerical_studies} and \ref{sec:conclusion}, respectively.


\subsection{Related Work}\label{sec:relatedWork}

There exists a substantial body of work on data-driven formal approaches for safety verification and control synthesis of stochastic systems.
We will focus on ``abstraction-free'' approaches via CBCs/CBFs, and refer to the complimentary publications by \shortciteA{gracia2023distributionally,banse2023data2,Chekan2023UncertainConstraints,makdesi2023data,schon2024btgp,kazemi2022datadriven,zhang2024formal,schoen2023bayesian,nazeri2025data} for a recent selection of abstraction-based work addressing uncertain systems.
  
%
For data-driven CBCs/CBFs, many existing approaches are limited to linear or control-affine dynamics~\shortcite<see, e.g.,>{Jagtap2020CBCGP,Cohen2022,Lopez2022uCBF}.
Furthermore, many approaches rely on known Lipschitz constants to provide formal guarantees and/or address only partially unknown dynamics.
For instance, Gaussian processes are employed by \shortciteA{Wang2018CBF} and by \citeA{Jagtap2020CBCGP} to learn partially unknown dynamics of nonlinear systems whilst assuming the affine control-dependent part of the dynamics to be known.
Systems with unknown additive disturbance are addressed by \citeA{Chekan2023UncertainConstraints}.
Similarly, \shortciteA{mathiesen2024data} assume the deterministic part of the dynamics to be accurately known, whilst the CBC-based safety control method proposed by \shortciteA{mazouz2024data} focus on known noise distributions instead.
Approaches to fully unknown dynamics are scarce. \shortciteA{Salamati2021DDCBC} study the computation of CBCs for fully unknown discrete-time systems relying on known Lipschitz constants.
Uncertain continuous-time systems are studied by \shortciteA{wang2023stochastic}, employing Bayesian inference and local Lipschitzness.

In the pursuit of a more flexible method, NN-based approaches for synthesizing so-called neural barriers have gained popularity due to their functional expressiveness \shortcite{so2023train,abate2021fossil,Safe_Barrier_Neural}.
For instance, \shortciteA{so2023train} train neural CBFs based on finite sample sets, acknowledging that their verification can be performed using NN verification tools such as the bound propagation techniques leveraged by \shortciteA{wang2024simultaneous}.
The field of NN verification is highly active, with numerous tools and methodologies being developed to ensure reliability and safety in AI systems \shortcite<see, e.g.,>{xu2020automatic,katz2017reluplex,kouvaros2021towards,henriksen2021deepsplit}. Neural networks as compact representations have been utilized by \shortciteA{majumdar2023neural} for memory-efficient formal verification and synthesis. 
Noteworthy adjacent work by \shortciteA{kazemi2020fullLTL} and \shortciteA{kazemi2024assume} studies temporal logic control via model-free reinforcement learning with convergence guarantees and assume guarantee contracts.
Apart from our previous work \cite{schon2024DRObarrier}, the only known publication on correct-by-design control via CMEs is due to \shortciteA{Romao2023DRControl}, which embeds the transition kernel for abstraction-based control. In contrast, we consider CMEs in an abstraction-free setting.
In extension to our previous work, \shortciteA{chen2025distributionally} propose a CBC-based approach for reach-avoid properties based on the Wasserstein distance.

\section{Preliminaries and Problem Statement}\label{sec:preAndProblem}
\noindent\textbf{Notation.}
We denote the sets of positive integers and non-negative reals as $\N_{>0}$ and $\R_{\geq 0}$, respectively.
Consider a Polish sample space $\X$ \cite{bogachev2007measure}.
Let $(\X,\borel{\X},\P)$ be the underlying probability space equipped with a Borel $\sigma$-algebra $\borel{\X}$ defined over $\X$, 
and a probability measure $\P$. 
For a random variable $X$, let $p_X$ be the pushforward probability measure of $\P$ under $X$ such that $X\sim p_X(\cdotx)$.
The expected value of a function $f(X)$ on $\X$ is written as $\E_{p_X}[f(X)]$. If it is clear from the context, we abbreviate and write $\E[f(X)]$.
We denote the set of all probability measures for a given measurable space $(\X,\borel{\X})$ as $\mathcal{P}(\X)$.
The $n$-dimensional Gaussian measure with mean $\mu\in\mathbb{R}^n$ and covariance matrix $\Sigma\in\mathbb{R}^{n\times n}$ is given by
\begin{equation*}
	\mathcal N(A\,|\,\mu, \Sigma) := {\int_A}\,\frac{{d\lambda^n(x)}}{ \sqrt{(2\pi)^n \left|\Sigma\right| }}\exp\left[-\frac{1}{2}(x-\mu)\T\Sigma^{-1\!}\,(x-\mu)\right],
\end{equation*}
where {$\lambda^n\colon\borel{\X}\to[0,\infty]$ denotes the usual $n$-dimensional Lebesgue measure and} $|\Sigma|$ is the determinant of $\Sigma$.
The Dirac delta measure $\delta_a\colon\borel{\X}\rightarrow [0,1]$ concentrated at a point $a\in\X$ is defined as $\delta_a(A)=1$ if $a\in A$ and $\delta_a(A)=0$ otherwise, for any measurable set $A\in\borel{\X}$.
We denote the uniform distribution over $\X$ as $\mathcal{U}_\X$ with realizations $x\sim\mathcal{U}_\X(\cdotx)$.
For two measurable spaces $(\X,\borel{\X})$ and $(\Y,\borel{\Y})$, a \emph{probability kernel} is a mapping $\p\colon \X \times \borel{\Y}\rightarrow  [0,1]$ such that $\p(X=x,\cdotx)\colon\borel{\Y}\rightarrow[0,1]$ is a probability measure for all $x\in\X$, and $\p(\cdotx, B)\colon \X\rightarrow [0,1]$ is measurable for all  $B\in\borel{\Y}$.
A probability kernel associates to each point $x\in\X$ a measure denoted by $\p(\cdotx|X=x)$.

The transpose of a vector or matrix $A$ is indicated as $A\T$. 
Let the $N\times N$ dimensional identity matrix be given by $I_N$.
Let $X_N:=[x_i]_{i=1}^N$ be a column vector with $x_i\in\X$.
We denote the element-wise evaluation of a function $f\colon\X\rightarrow\R$ on $X_N$ as $f(X_N):=[f(x_i)]_{i=1}^N$.
Similarly, we may write $A = [a_{ij}]_{i,j=1}^N$ to denote a matrix with its elements.
{For a vector $v$, we denote by $v^{-1}$ its element-wise inverse.}

\subsection{Discrete-Time Stochastic Systems}
In this work, we consider systems expressible as Markov decision processes over continuous state and input spaces, formally defined as follows.
\begin{definition}[Markov decision process (MDP)]
    An MDP is a tuple $\M=(\X,\X_0,\U,\Tr)$, comprising a state space $\X\subset\R^n$ with states $x\in\X$; initial states $x_0\in \X_0\subset\X$; an input space $\U$ with inputs $u\in\U$; and a probability kernel $\Tr\colon\X\times\U\times\borel{\X}\rightarrow[0,1]$.
\end{definition}

In every execution, given a current state $x\in\X$ and control input $u\in\U$, the MDP evolves to a consecutive state $x_{+}\in\X$, which is obtained as a realization $x_{+}\sim\Tr(\,\cdot\,|X=x,U=u)$.
As a notable class of systems that can be captured using MDPs, we may consider black-box systems with Markovian discrete-time stochastic dynamics, namely
\begin{equation}
	\label{eq:model}
	\M\colon \left\{ \begin{array}{ll}
		x_{t+1}= f(x_t,u_t,w_t),\quad w_t\sim p_w(\cdotx),\\
	\end{array} \right.
\end{equation}
where the system state and control input at the $t^{\text{th}}$ time-step are denoted by $x_t$ and $u_t$, respectively.
The state evolution of the system, described by the function $f\colon\X\times\U\times\mathbb{W}\rightarrow\X$,
is subject to {independent, identically distributed (i.i.d.) noise} $w_t\sim p_w(\cdotx)$ supported on a set $\mathbb{W}$. 
{The corresponding probability kernel is given by
\begin{equation*}
    \Tr(dx_+|x_t,u_t)= \int_{\mathbb{W}}\,\delta_{f(x_t,u_t,w)}(dx_+)\,p_w(dw).
\end{equation*}
}

\subsection{System Safety}
In this paper, we {aim} to synthesize a {control policy} $\pi$ such that the {resulting} closed-loop system $\pi\times\M$ is safe.
{We characterize \emph{safety} associated with a time horizon $T \in\N_{>0}$ as a property of state trajectories: a trajectory is said to be safe if it does not enter a designated unsafe set $\X_u\subset\X$ for all $t \in [0, T]$. 
As stochastic systems do generally not admit a binary notion of safety, this induces a safety probability on the underlying closed-loop system, defined as the probability that the trajectory of $\pi\times\M$ starting from any initial state $x_0\in \X_0$ remains in the safe set $\X\setminus\X_u$.
In the following, we focus on a finite-horizon formulation.}
Thus, we quantify the \emph{probability} of reaching $\X_u$ in a given finite horizon $T < \infty$ and compute its complement.
A (finite-horizon) \emph{safety specification} $\psi_{\mathrm{safe}}:=(\X_u,T)$ is hence fully characterized by an unsafe set $\X_u$ and a time horizon $T$.
The system $\pi\times\M$ satisfies $\psi_{\mathrm{safe}}$ with probability at least $p\in(0,1)$ if the probability of its trajectories starting in $\X_0$ and avoiding $\X_u$ within horizon $T$ is at least $p$. This is denoted by $P^\pi_{\mathrm{safe}}(\M):=\P(\pi\times\M\satisfies\psi_{\mathrm{safe}})$~\cite{BK08}, {for which we require $P^\pi_{\mathrm{safe}}(\M)\geq p$.}

{Consider the following concrete example.}

\begin{figure}
    \centering
    \def\svgwidth{\linewidth}
\begingroup%
  \makeatletter%
  \providecommand\color[2][]{%
    \errmessage{(Inkscape) Color is used for the text in Inkscape, but the package 'color.sty' is not loaded}%
    \renewcommand\color[2][]{}%
  }%
  \providecommand\transparent[1]{%
    \errmessage{(Inkscape) Transparency is used (non-zero) for the text in Inkscape, but the package 'transparent.sty' is not loaded}%
    \renewcommand\transparent[1]{}%
  }%
  \providecommand\rotatebox[2]{#2}%
  \newcommand*\fsize{\dimexpr\f@size pt\relax}%
  \newcommand*\lineheight[1]{\fontsize{\fsize}{#1\fsize}\selectfont}%
  \ifx\svgwidth\undefined%
    \setlength{\unitlength}{515.90551181bp}%
    \ifx\svgscale\undefined%
      \relax%
    \else%
      \setlength{\unitlength}{\unitlength * \real{\svgscale}}%
    \fi%
  \else%
    \setlength{\unitlength}{\svgwidth}%
  \fi%
  \global\let\svgwidth\undefined%
  \global\let\svgscale\undefined%
  \makeatother%
  \begin{picture}(1,0.54945055)%
    \lineheight{1}%
    \setlength\tabcolsep{0pt}%
    \put(0,0){\includegraphics[width=\unitlength,page=1]{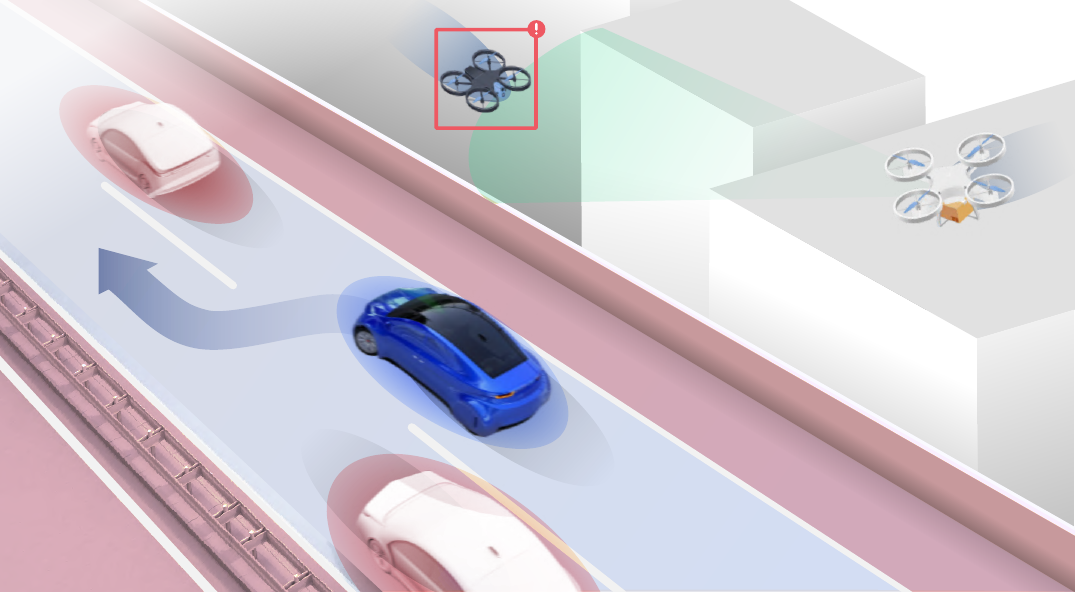}}%
    \put(0.37622381,0.182704){\color[rgb]{0,0.21568627,1}\makebox(0,0)[t]{\lineheight{1.25}\smash{\begin{tabular}[t]{c}$\X_0$\end{tabular}}}}%
    \put(0.6487633,0.16941234){\color[rgb]{0.61568627,0,0}\transparent{0.90118003}\makebox(0,0)[t]{\lineheight{1.25}\smash{\begin{tabular}[t]{c}$\X_u$\end{tabular}}}}%
    \put(0.05950908,0.06116456){\color[rgb]{0.61568627,0,0}\transparent{0.90118003}\makebox(0,0)[t]{\lineheight{1.25}\smash{\begin{tabular}[t]{c}$\X_u$\end{tabular}}}}%
  \end{picture}%
\endgroup%

    \caption{Examples of safety-critical embodied AI systems in the transport sector.}
    \label{fig:safety_setup}
\end{figure}
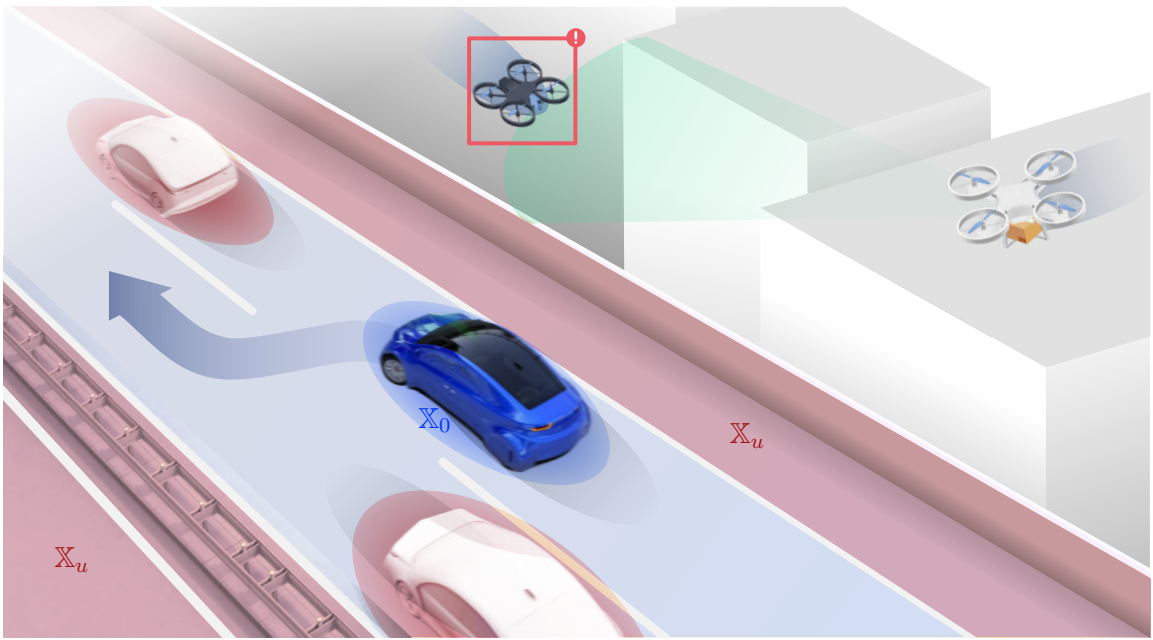

{\begin{example}\label{exmp:overtaking}
    Figure~\ref{fig:safety_setup} shows an autonomous vehicle (in blue) overtaking a leading vehicle. 
    The dynamics of the ego vehicle is given by Dubin's car model with additive noise:
    \begin{align}
    	\begin{bmatrix} {x}_{t+1}\\{y}_{t+1}\\\phi_{t+1} \end{bmatrix} = \begin{bmatrix} {x}_{t}\\{y}_{t}\\\phi_{t} \end{bmatrix} + 
    	\tau \begin{bmatrix} 
    		v\cos(\phi_t)\\
    		v \sin(\phi_t)\\
    		u_t
    	\end{bmatrix} + \begin{bmatrix} w^1_t\\w^2_t\\w^3_t  \end{bmatrix},\label{eq:car}
    \end{align}
    {with time discretization $\tau:=0.5$,}
    where $(x,y)$ denotes the relative position in longitudinal and lateral direction, and $\phi$ is the heading angle. 
    The maneuver starts in $\X_0$ (blue region).
    Let the noise $w_t^i$, $i\in\{1,2,3\}$, be zero-mean Gaussian with standard deviation $0.01$, $0.01$, and $0.001$, respectively.
    The steering wheel angle $u_t=\pi(x_t)$ is supplied by a 
    neural network (NN) controller $\pi\colon\X\rightarrow\U$, which is trained to steer clear from the sides of the road and overtake the leading vehicle without collision, that is, as illustrated in Figure~\ref{fig:safety_setup}, avoiding the unsafe regions $\X_u$ and reaching a target set ahead of the leading vehicle.
    The goal of this paper is to determine (probabilistically) if the controlled system $\pi\times\M$ is safe.
\end{example}}

{
We remark that for finite-horizon safety, maximizing $P^\pi_{\mathrm{safe}}(\M)$ with respect to the policy $\pi$ would require policies that are Markov and time-variant \cite{puterman1994MDPs}. In contrast, the inequality $P^\pi_{\mathrm{safe}}(\M)\ge p$ may be satisfiable by any type of policies. For simplicity, we consider stationary (i.e., time-invariant) policies of the form $\pi:\X\rightarrow \U$.
}

\subsection{Control Barrier Certificates}\label{sec:cbc}
Whilst certifying the safety of a {continuous-space} stochastic system is generally challenging, \emph{control barrier certificates} (CBCs) and \emph{control barrier functions} (CBFs) leverage the concept of set invariance to arrive at an abstraction-free formulation. This has made CBCs/CBFs popular tools for safety verification and synthesis~{\cite{prajna2007framework}}.
We briefly recall the definition of a CBC{, based on the theory of \citeA{kushner1967stochastic}, dubbed \emph{stochastic barrier functions} by \citeA{santoyo2021barrier}}.
\begin{definition}[Control barrier certificate (CBC)]\label{def:cbc}
    A function $\B\colon\X\rightarrow\R_{\geq 0}$ is called a CBC of an MDP $\M=(\X,\X_0,\U,\Tr)$ with reference to an unsafe set $\X_u$, if we have
    \begin{itemize}
        \item[(a)] $\forall x_0\in \X_0\colon\,\B(x_0)\leq\eta$;
	\item[(b)] $\forall x_u\in \X_u\colon\,\B(x_u)\geq{1}$; and
	\item[(c)] $\forall x\in\X,\exists u\in\U\colon\,\E_{\Tr}[\,\B(X^+)\mid X=x,\,U=u\,] -\B(x) \leq  c;$
    \end{itemize}
    for some constants ${\eta\in[0,1)}$ and $c\geq 0$.
\end{definition}
Intuitively, condition (c) of Definition~\ref{def:cbc} restricts the CBC $\B$ to elicit a relaxed \emph{supermartingale} property for any 
non-zero constant $c$.
Note that this implies the existence of a stationary policy $\pi\colon\X\rightarrow\U$ generating the corresponding control inputs $u\in\U$.
{If a CBC can be found} for a system $\M$, then, a lower bound on the probability of $\M$ being safe is given by the following proposition due to \citeA[Theorem~3]{kushner1967stochastic}.
\begin{proposition}[Finite-horizon safety]\label{prop:reach}
    Consider an MDP $\M=(\X,\X_0,\U,\Tr)$ and a safety specification $\psi_{\mathrm{safe}}=(\X_u,T)$. Suppose there exists a CBC $B$ w.r.t. $\X_u$ (Definition~\ref{def:cbc}) with constants $\eta$ and $c$. Then, there exists a stationary policy $\pi$ such that
    $$P^\pi_{\mathrm{safe}}(\M)\geq 1- {(\eta + cT)}.$$
\end{proposition}
\begin{remark}[Infinite-horizon safety]\label{rem:ininite_horizon_safety}
    Finding a CBC (Definition~\ref{def:cbc}) for a non-negative constant $c>0$ is generally easier than for $c=0$. If, however, a CBC can be obtained for $c=0$, Proposition~\ref{prop:reach} provides safety guarantees for an unbounded time horizon $T\rightarrow\infty$.
    {This recovers the results by \shortciteA{prajna2007framework}.}
\end{remark}

\subsection{Problem Statement}
{The aim is to synthesize a safety} policy $\pi$ for a black-box system $\M$, {where} the transition kernel $\Tr$ is unknown and only observed through a finite set of observations of the form $\{x^i,u^i,x^i_+\}_{i=1}^N${, $N\in\N_{>0}$}, where $x^i_{+} \sim \Tr(\cdotx\mid X=x^i,\,U=u^i)$ for uniformly drawn $x^i\sim\mathcal{U}_\X(\cdotx)$ and $u^i\sim\mathcal{U}_\U(\cdotx)$.
This corresponds to drawing i.i.d. samples from the joint distribution 
\begin{equation}
    (X,U,X^+)\sim\Tr(X^+|X,U)\,\mathcal{U}_\X(X)\,\mathcal{U}_\U(U).\label{eq:data_generation}
\end{equation}
Note that this setting admits a plethora of target systems.
We formalize the problem statement as follows.

\begin{problem}\label{prob:main}
    Consider a given safety specification $\psi_{\mathrm{safe}}$ and a confidence \emph{level} $1-\rho\in(0,1)$. Without knowledge of the transition kernel $\Tr$ of $\M$ and based only on i.i.d. observations \mbox{$\{x^i,u^i,x^i_+\}_{i=1}^N$, $N\in\N_{>0}$,} find a policy $\pi\colon\X\rightarrow\U$ and threshold $p^\pi_N\in (0,1)$ such that $P^\pi_{\mathrm{safe}}(\M) \geq p^\pi_N$ with {confidence} at least $1-\rho$.
\end{problem}

{The confidence in Problem~\ref{prob:main} is with respect to a random draw of $N$ data samples used to obtain the safety bound $p^\pi_N$, i.e., we have $\P^N(P^\pi_{\mathrm{safe}}(\M) \geq p^\pi_N)\geq 1-\rho$, where $\P^N$ denotes the joint probability distribution over $N$ data samples.}
    
In this work, we address Problem~\ref{prob:main} under the assumption of minimal complexity information (see Assumption~\ref{asm:BCinRKHS}) and use CBCs to certify probabilistic safety. 
The main challenge in establishing CBCs for unknown stochastic systems arises from the {stochastic} constraint in condition (c) (Definition~\ref{def:cbc}). We show how CBCs can be generated from data by embedding the conditional probability measure via  kernel methods, which cast the {stochastic} constraint {in terms of an} inner product 
We construct an RKHS ambiguity set that is centered at the empirical mean embedding and can be inflated to robustify against out-of-distribution dynamics.
As a result, the problem of synthesizing a robust policy is cast as a semi-infinite program.
For the special case where a policy $\pi$ is given, i.e., system verification, we present an efficient algorithmic solution.
Based on a finite Fourier expansion of the kernel, we establish a relaxation of the semi-infinite program as a scalable linear program.

\smallskip

In the following section, we give a brief introduction to RKHS theory and provide the results for embedding (conditional) probability measures into RKHSs.

\section{Kernel Mean Embeddings}\label{sec:RKHS}
\noindent\textbf{RKHS basics.}
A symmetric function $k_\X\colon\X\times\X\rightarrow\R$ is called a (positive definite) \emph{kernel} (note the distinction from \emph{probability kernels}) if for all $N\in\N_{>0}$ we have $\sum_{i=1}^{N}\sum_{j=1}^{N}a_i a_j\allowbreak k_\X(x_i,x_j) \geq 0$ for $x_1,\ldots,x_N\in\X\subset{\R^n}$ and $a_1,\ldots,a_N\in\R$.
A prominent example is the \emph{squared exponential} (SQExp) kernel \shortcite{Rasmussen2005GP,Kanagawa2018GPvsKernel}:
\begin{equation}
    k_\X(x,x') := \sigma_f^2 \exp\left( -\frac{1}{2} (x-x')\T \Sigma^{{-1}} (x-x') \right),\quad \Sigma:=\diag(\sigma_l)^{{2}},\label{eq:sqexp_kernel}
\end{equation}
with amplitude $\sigma_f^2\geq0$ and lengthscale coefficients $\sigma_l\in\R^n$.
For this work, we assume that all kernels are bounded on their domain, i.e., $\E_{}[k_\X(x,x)]<\infty$, $x\in\X$.
Given a kernel $k_\X$ on a non-empty set $\X$, there exists a unique corresponding
\emph{reproducing kernel Hilbert space} (RKHS) $\Hilbert_{k_\X}$
of functions $f\colon\X\rightarrow\R$ equipped with an inner product $\innerH{\cdotx}{\cdotx}{\Hilbert_{k_\X}}$
with the {celebrated} \emph{reproducing property} such that for any function $f\in\Hilbert_{k_\X}$ and $x\in\X$ we have $f(x)=\innerH{f}{k_\X(\cdotx,x)}{\Hilbert_{k_\X}}$.
Note that {$\phi_\X:=k_\X(\cdotx,x)\colon\X\rightarrow\Hilbert_{k_\X}$} is a real-valued function,
which is also called an implicit \emph{canonical} \emph{embedding} or \emph{feature map} 
such that $k_\X(x,{x'})=\innerH{\phi_\X(x)}{\phi_\X(x')}{\Hilbert_{k_\X}}$ for all $x,x'\in\X$.
For an RKHS $\Hilbert_{k_\X}$, we use the associated feature map $\phi_\X$ and kernel $k_\X$ interchangeably for ease of notation and comprehensibility.
The inner product induces the norm $\norm{f}_{\Hilbert_{k_\X}}\!\!\!\!:=\!\!\sqrt{\smash[b]{\innerH{f}{f}{\Hilbert_{k_\X}}}}$ of the RKHS. 
{A Hilbert space $\Hilbert$ is said to be separable if it admits a countable dense subset.}
Throughout this paper, we assume that all RKHSs are \emph{separable}.
Refer to the monograph by \citeA{Berlinet2004RKHSProbStat} for a comprehensive study on RKHSs.
Given $N$ i.i.d. samples $\hat X_N:=[\hat{x}_i]_{i=1}^N$ with $\hat{x}_i\in\X$, the
\emph{Gram matrix} of $k_\X$ is given by
$K_{\hat{X}}^N:=[k_\X(\hat{x}_i,\hat{x}_j)]_{i,j=1}^N.$
Furthermore, we define the vector-valued function 
$k_{\hat{X}}^N(x)  := [k_\X(x,\hat{x}_i)]_{i=1}^N.$

\medskip

\noindent\textbf{Tensor product Hilbert spaces.} For two kernels $k_\X\colon\X{\times\X}\rightarrow\R$ and $k_\Y\colon\Y{\times\Y}\rightarrow\R$ with associated RKHSs $\Hilbert_\X$ and $\Hilbert_\Y$,
$\Hilbert_{\X\Y}:=\Hilbert_{\X}\otimes\Hilbert_{\Y}$ is the \emph{tensor product of the Hilbert spaces $\Hilbert_{\X}$ and $\Hilbert_{\Y}$} with reproducing kernel $k_{\X\Y}((x,y),(x',y'))=k_\X(x,x')\,k_\Y(y,y')$, for $x,x'\in\X$, $y,y'\in\Y$. $\Hilbert_{\X}\otimes\Hilbert_{\Y}$ is equipped with the inner product $\innerH{{\varphi_\X}\otimes{\varphi_\Y}}{{\varphi_\X}'\otimes{\varphi_\Y}'}{\Hilbert_{\X}\otimes\Hilbert_{\Y}}=\innerH{{\varphi_\X}}{{\varphi_\X}'}{\Hilbert_{\X}}\,\innerH{{\varphi_\Y}}{{\varphi_\Y}'}{\Hilbert_{\Y}}$, for ${\varphi_\X},{\varphi_\X}'\in\Hilbert_\X$ and ${\varphi_\Y},{\varphi_\Y}'\in\Hilbert_\Y$, where its elements ${\varphi_\X}\otimes{\varphi_\Y}$ are called \emph{tensors}. {The associated norm is given by} $\norm{{\varphi_\X}\otimes{\varphi_\Y}}_{\Hilbert_{\X}\otimes\Hilbert_{\Y}}=\norm{{\varphi_\X}}_{\Hilbert_{\X}}\norm{{\varphi_\Y}}_{\Hilbert_{\Y}}$ \cite[Chapter~2]{arai2018analysis}.
For feature vectors ${\varphi_\X}(x)$ and ${\varphi_\Y}(y)$, ${\varphi_\X}(x)\otimes{\varphi_\Y}(y)$ indicates their outer product.

\subsection{Embedding Probability Measures}
To reason about the expected value of a random variable, embedding the variable into a (higher dimensional) space is a well-established concept in ML \cite{scholkopf2002learning,Steinwart2008SVM}.
The \emph{(kernel) mean embedding} (ME) follows the same reasoning and represents the projection of a probability measure into an RKHS~\shortcite{Smola2007EmbedDistrb}.
\begin{definition}[Mean embedding (ME)]
    Given an RKHS $\Hilbert_{k_\X}$ {induced} by a kernel $k_\X\colon\X\times\X\rightarrow\R$, the \emph{mean embedding} of a probability measure $p\colon\borel{\X}\rightarrow[0,1]$ is computed via the \emph{mean map} $\mu_{k_\X}\colon\mathcal{P}(\X)\rightarrow\Hilbert_{k_\X}$ as
    \begin{equation*}
	\mu_{k_\X}(p) := \E_{p}[\phi_\X(X)] = \int_{\X}\phi_\X(x)\,dp(x).
    \end{equation*}
\end{definition}
Note that the reproducing property of the kernel carries on to the ME, facilitating the computation of the expected value of a function $f\in\Hilbert_{k_\X}$ via the inner product \cite{Smola2007EmbedDistrb}, that is
\begin{equation}
	\E_{p}[f(X)] = \innerH{f}{{\mu}_{k_\X}(p)}{\Hilbert_{k_\X}}.\label{eq:innerProdMean}
\end{equation}
If $k_\X$ is \emph{characteristic}, any probability measure $p\in\mathcal{P}(\X)$ is injectively mapped to a unique ME $\mu_{k_\X}(p)\in\Hilbert_{k_\X}$ \shortcite{Gretton2012}.
This gives rise to the definition of a distance metric between {two probability measures in terms of their} embeddings in RKHS --- the \emph{maximum mean discrepancy} (MMD). For a characteristic kernel $k_\X$, the MMD between two probability measures $p,p'\in\mathcal{P}(\X)$ in $\Hilbert_{k_\X}$ is defined as $\norm{\mu_{k_\X}(p)-\mu_{k_\X}(p')}_{\Hilbert_{k_\X}}$ \cite{Gretton2012}.

\subsection{Embedding Conditional Probability Measures}
Analogous to generic probability measures, there exists a similar notion for embedding \emph{conditional} probability measures of the form $\p\colon\X\times\borel{\Y}\rightarrow[0,1]$ with realizations $Y\sim \p(\cdotx|X=x)$ for a conditioning variable taking concrete values $x\in\X$. 
More specifically, we use the measure-theoretic \emph{conditional mean embedding} (CME) introduced by \citeA{Park2020MeasureTheoretic}. 
To this end, we equip the space of the conditioning random variable $X\in\X$ and the space of the target random variable $Y\in\Y$ with individual kernels $k_\X\colon\X\times\X\rightarrow\R$ and $k_\Y\colon\Y\times\Y\rightarrow\R$, respectively.
\begin{definition}[Conditional mean embedding (CME)]\label{def:condMeanEmbed}
    Given two RKHSs $\Hilbert_{k_\X}$ and $\Hilbert_{k_\Y}$ with the associated kernels $k_\X$ and $k_\Y$, the \emph{CME} of a conditional probability measure $\p\colon\X\times\borel{\Y}\rightarrow[0,1]$ is an $X$-measurable random variable taking values in $\Hilbert_{k_\Y}$ given by
    \begin{equation*}
	\mu_{k_\Y|k_\X}(\p)(\cdotx) := \E_{\p}[\phi_\Y(Y)\mid X=\cdotx].
    \end{equation*}
\end{definition}
Refer to \citeA{Park2020MeasureTheoretic} for a comprehensive mathematical dissemination of the CME.
Analogous to the non-conditional case, we can compute the conditional expectation of a function $f\in\Hilbert_{k_\Y}$ via its inner product with the CME, i.e., almost surely
\begin{equation}
	\E_{\p}[f(Y)\mid X=x] = \innerH{f}{{\mu}_{k_\Y|k_\X}(\p)(x)}{\Hilbert_{k_\Y}}.\label{eq:innerProdCond}
\end{equation}
Since the CME of a conditional measure $\p$ is generally unknown (and potentially infinite dimensional), we may obtain an empirical estimate from a finite set of training data.
\begin{proposition}[Empirical CME]\label{prop:empCondMeanEmbed}
    Let a finite dataset $(\hat{X}_N,\hat{Y}_N):=([\hat{x}_i]_{i=1}^N,[\hat{y}_i]_{i=1}^N)$ of samples $(\hat{x}_i,\hat{y}_i)\in\X\times\Y$ from a conditional prob. measure $\p\colon\X\times\borel{\Y}\rightarrow[0,1]$ be given s.t. $\hat{y}_i\sim \p(\cdotx|X=\hat{x}_i)$.
    For kernels $k_\X$, $k_\Y$, the \emph{empirical CME} of $\p$ given $(\hat{X}_N,\hat{Y}_N)$ is
    \begin{equation*}
    	\hat{\mu}_{k_\Y|k_\X}^N(\cdotx) := k_{\hat{X}}(\cdotx)\T\left[ K_{\hat{X}}^N+N\lambda I_N\right]^{-1\!} \phi_\Y(\hat{Y}_N),
    \end{equation*}
    with a regularization constant $\lambda\geq0$.
    For $N\rightarrow\infty$, the empirical CME converges in expectation, that is, $\E_{\mathcal{U}_\X\mathcal{U}_\U}\big[\big\vert\big\vert\hat{\mu}_{k_\Y|k_\X}^N(X,U)-{\mu}_{k_\Y|k_\X}(\p)(X,U)\big\vert\big\vert_{\Hilbert_{k_\Y}}\big]\rightarrow0$.
\end{proposition}

Here and in the following, we assume that $[K_{\hat{X}}^N+N\lambda I_N]$ (and equivalent terms) are invertible. Note that this is always true if the regularization constant $\lambda$ is strictly positive.
By virtue of the reproducing property we have for any function $f\in\Hilbert_{k_\Y}$ almost surely that 
\begin{equation}
	\E_{\p}[f(Y)\mid X=x] \approx \innerH{f}{\hat{\mu}^N_{k_\Y|k_\X}(x)}{\Hilbert_{k_\Y}} = k_{\hat{X}}^N(x)\T\left[ k_{\hat{X}}^N+N \lambda I_N\right]^{-1\!} f(\hat{Y}_N).\label{eq:empiricalInnerProd}
\end{equation}
Note that the empirical estimate improves in probability as more data becomes available.

\section{Data-Driven Barrier Formulation}
\label{sec:ddcbc}
In this section, we establish the supermartingale-like property (c) in Definition~\ref{def:cbc} for a system with unknown transition kernel $\Tr$ using its CME based on i.i.d. training data generated from the unknown true system \eqref{eq:data_generation}, that is, data of the form
\begin{align}
    (\hat{X}_N, \hat{U}_N, \hat{X}^{+\!}_N) := ([\hat{x}_i]_{i=1}^N, [\hat{u}_i]_{i=1}^N, [\hat{x}^+_1]_{i=1}^N), \quad \text{with} \quad \hat{x}^+_i\sim\Tr(\cdotx|X=\hat{x}_i,\,U=\hat{u}_i),
    \label{eq:data}
\end{align}
generated from the unknown black-box system as shown in \eqref{eq:data_generation}.

{Our theoretical analysis assumes uniform i.i.d. samples to simplify concentration arguments --- a common assumption in formal settings.
In particular, this assumption stems from established results \cite<e.g.,>{Park2020MeasureTheoretic} on the consistency of the empirical CME (see Proposition~\ref{prop:empCondMeanEmbed}).
Whilst access to i.i.d. data of the form \eqref{eq:data} is a strong assumption, the guarantees established in this manuscript can be preserved under substantially weaker conditions, such as ergodicity, geometric mixing, or risk-based notions, by replacing the nominal sample size $N$ with an effective sample size $N_{\text{eff}}$ that accounts for temporal dependence \cite<see, e.g.,>{massiani2024consistency,steinwart2009fast,ziemann2022learning}.
Whilst the development of new concentration inequalities lies beyond the scope of this paper, it has been shown that kernel-based methods such as the CME can be applied even with dependent data --- such as data sequences --- in many settings \cite<see, e.g.,>{massiani2024consistency,steinwart2009fast,ziemann2022learning}.}

In the following, we will use three kernels, one for each of the three components $\hat{X}_N$, $\hat{U}_N$, and $\hat{X}^{+\!}_N$, namely
\begin{equation}
			k_{x}\colon\X\times\X\rightarrow\R,\quad
                k_{u}\colon\U\times\U\rightarrow\R,\quad
                k_{+\!}\colon\X\times\X\rightarrow\R,
			\label{eq:kernels}
\end{equation}
with their associated RKHSs $\Hilbert_{x}$, $\Hilbert_{u}$, and $\Hilbert_{+}$, respectively. We will assume that $k_{+\!}$ is characteristic.
Furthermore, we define the tensor product space $\Hilbert_{xu}:=\Hilbert_x\otimes\Hilbert_u$ with the associated kernel $k_{xu}:=k_x\cdot k_u$.
{Table~\ref{tbl:rkhs_overview} gives a brief summary of the notation used in the remainder of the paper.}
\begin{table}
    
    \centering
    \begin{tabular}{ccc}
        \toprule
        \textbf{RKHS}          & \textbf{Kernel}                                                                    & \textbf{Feature Map}                     \\
        \midrule
        $\Hilbert_x$    & $k_x\colon\X\times\X\rightarrow\R$                                          & $\phi_x\colon\X\to\Hilbert_x$     \\
        $\Hilbert_u$    & $k_u\colon\U\times\U\rightarrow\R$                                          & $\phi_u\colon\U\to\Hilbert_u$     \\
        $\Hilbert_+$    & $k_{+}\colon\X\times\X\rightarrow\R$                                        & $\phi_{+}\colon\X\to\Hilbert_{+}$ \\
        $\Hilbert_{xu}$ & $k_{xu}\colon(\X\times\U)\times(\X\times\U)\rightarrow\R$                   & not used                          \\
        $\mathcal{G}$      & $\Gamma\colon(\X\times\U)\times(\X\times\U)\rightarrow\mathcal{L}(\Hilbert_{+})$  & not used                          \\
        \bottomrule
    \end{tabular}
    \caption{{Overview of the kernels and RKHSs.}}
    \label{tbl:rkhs_overview}
\end{table}
To reason about the unknown dynamics in a mathematically rigorous way, we must restrict the system dynamics to a known function space.
In particular, we will assume that the CME of the transition kernel $\Tr$ lives in a vector-valued RKHS of mappings $\X\times\U\rightarrow\Hilbert_{+}$, which we indicate by $\mathcal{G}$ \cite[Section~2.3]{Park2020MeasureTheoretic}.\footnote{For a fixed choice of $(x,u)\in\X\times\U$, the vector-valued RKHS $\mathcal{G}$ reduces to the regular RKHS $\Hilbert_+$.}
We will call $\mathcal{G}$ the latent \emph{hypothesis (function) space} of the system.
Under standard assumptions, $\mathcal{G}$ is dense in $L_2(\X\times\U,\borel{\X\times\U},\Tr;\Hilbert_+)$, where $L_2(\X\times\U,\borel{\X\times\U},\Tr;\Hilbert_+)$ is the space of strongly $(\borel{\X\times\U}-\borel{\Hilbert_+})$-measurable and Bochner square-integrable functions $\X\times\U\rightarrow\Hilbert_+$ w.r.t. $\Tr$, with $\borel{\X\times\U}$ and $\borel{\Hilbert_+}$ being the Borel $\sigma$-fields of $\X\times\U$ and $\Hilbert_+$, respectively \cite{li2022optimal}.
{This assumption imposes minimal structural constraints on the underlying dynamics.
In fact, in contrast to many existing approaches based on finite Taylor approximations of the unknown dynamics --- which are limited to approximating analytic functions of bounded degree --- the RKHS framework adopted here allows approximation of a far broader class, e.g., of all smooth functions via the squared exponential kernel.
This property affords significantly greater modeling flexibility and expressiveness. Moreover, the choice of kernel provides a natural mechanism for incorporating prior knowledge, where available.}
With this, we raise the following standard assumption.

\begin{assumption}\label{asm:BCinRKHS}
    Let the CME $\mu_{k_{+\!}|k_{xu}}(\Tr)$ be \emph{well-specified}, i.e.,  $\mu_{k_{+\!}|k_{xu}}(\Tr)\in\mathcal{G}$. Furthermore, given confidence \mbox{$1-\rho\in(0,1)$}, let a bound $\varepsilon\geq0$ be known such that $$\P\left(\left\vert\left\vert\mu_{k_{+\!}|k_{xu}}(\Tr)-\hat\mu^N_{k_{+\!}|k_{xu}}\right\vert\right\vert_{\mathcal{G}}\leq\varepsilon\right)\geq 1-\rho,$$ for the empirical CME based on $(\hat{X}_N, \hat{U}_N, \hat{X}^{+\!}_N)$, which is given by
    \setlength{\belowdisplayskip}{3pt}
    \begin{equation}
        \hat\mu^N_{k_{+\!}|k_{xu}} := k^N_{\hat{X}\hat{U}}(\cdotx)\T\left[ K^N_{\hat{X}\hat{U}}+N\lambda I_N\right]^{-1\!} \phi_+(\hat{X}^+_N).
        \label{eq:empiricalCME}
    \end{equation}
\end{assumption}

It is common practice to {introduce a robustness} radius $\varepsilon$ {(as in Assumption~\ref{asm:BCinRKHS})} to {account for uncertainty in empirical estimates} such as the estimated conditional expectation \eqref{eq:empiricalInnerProd} by constructing an RKHS ambiguity set centered at the empirical CME. 
{Although theoretical results exist that relate this radius to the probability of the true CME of the black-box dynamics $\Tr$ lying within the ambiguity set --- achieving minimax rates of order $\mathcal{O}(\log(N)/N)$~\cite{li2022optimal,mollenhauer2022learning} --- such concentration bounds tend to be excessively conservative in practice. Consequently, the radius is often chosen manually or calibrated using data-driven techniques such as MMD-based bootstrapping~\shortcite<see, e.g.,>[for an approach using non-conditional MEs]{Nemmour2022FiniteSampleGuarantee}.}

To learn CBCs from data, we start by reformulating the conditional expectation in the left-hand side of condition (c) in Definition~\ref{def:cbc} via the inner product with the CME $\mu_{k_{+\!}|k_{xu}}(\Tr)$ as shown in \eqref{eq:innerProdCond}, that is, almost surely
\begin{equation*}
	\E_{\Tr}[\B(X^+)|X=x,\,U=u] = \innerH{\B}{\mu_{k_{+\!}|k_{xu}}(\Tr)(x,u)}{\Hilbert_{+}}.
\end{equation*}
Since the CME $\mu_{k_{+\!}|k_{xu}}(\Tr)$ is unknown, we construct an ambiguity set $\ambRKHS^{N}_\varepsilon\subset\mathcal{G}$ centered at the empirical CME $\hat\mu^N_{k_{+\!}|k_{xu}}$ constructed from the data in \eqref{eq:data} (via Proposition~\ref{prop:empCondMeanEmbed}) and choose an MMD radius $\varepsilon\geq0$ such that the CME of $\Tr$ lies within $\ambRKHS^{N}_\varepsilon$ with a confidence of at least $1-\rho\in(0,1)$, i.e., such that $\P(\mu_{k_{+\!}|k_{xu}}(\Tr)\in\ambRKHS^{N}_\varepsilon)\geq 1-\rho$, with the ambiguity set
\begin{equation}\label{eq:ambRKHS}
	\ambRKHS^{N}_\varepsilon
	 :=  \left\lbrace\mu\in\mathcal{G} \left|\, \norm{\mu-\hat\mu^N_{k_{+\!}|k_{xu}}}_{\mathcal{G}} \leq \varepsilon\right.\right\rbrace.
\end{equation}
With this, we obtain the following result.
\begin{theorem}\label{thm:cbc1}
    Let data $(\hat{X}_N, \hat{U}_N, \hat{X}^{+\!}_N)$ in \eqref{eq:data} from an unknown MDP $\M$ and
    kernels $k_{xu}$ and $k_{+\!}$ (characteristic) be given.
    Consider the resulting empirical CME $\hat\mu^N_{k_{+\!}|k_{xu}}$ in \eqref{eq:empiricalCME} and the ambiguity set $\ambRKHS^{N}_\varepsilon$ in \eqref{eq:ambRKHS} with confidence bound $1-\rho$.
    If there exists	a function $\B\colon\X\rightarrow\R_{\geq 0}$, $\B\in\Hilbert_+$, satisfying
    \begin{equation}
	\forall x\in\X,\,\exists u\in\U,\,\forall \mu\in\ambRKHS^{N}_\varepsilon\colon \innerH{\B}{\mu(x,u)}{\Hilbert_{+}} -\B(x)  \leq c,\label{eq:ineqCondForallX}
    \end{equation}
	for some constant $c\geq 0$,
	then,
	$\B$ satisfies CBC condition (c) of Definition~\ref{def:cbc} w.r.t. $\M$ with probability at least $1-\rho$.
\end{theorem}

With the following theorem, we provide a way of establishing Theorem~\ref{thm:cbc1} by reformulating the left-hand side of \eqref{eq:ineqCondForallX} for the RKHS norm-ball ambiguity set in \eqref{eq:ambRKHS}.
For this, let $K_{\hat{X}^{+\!}}^N$ and $K_{\hat{X}\hat{U}}^N$ be the Gram matrices associated with $(k_{+\!},\hat{X}^+)$ and $(k_{xu},[(\hat{x}_i,\hat{u}_i)]_{i=1}^N)$, respectively.
\begin{theorem}\label{thm:cbc2}
    Consider the setup of Theorem~\ref{thm:cbc1}. If there exists a function $\B\colon\X\rightarrow\R_{\geq 0}$, $\B\in\Hilbert_+$, with some $\bar{\B}\geq\norm{B}_{\Hilbert_{+}}$ such that
    \begin{equation}
	\forall x\in\X,\,\exists u\in\U\colon w(x,u)\T \B(\hat{X}^+_N) -\B(x)\leq  c - \varepsilon\bar{\B}\kappa(x,u),\label{eq:ineqCondForallX2}
    \end{equation}
    for some constant $c\geq 0$, $\kappa(x,u):=\sqrt{k_x(x,x)}\sqrt{k_u(u,u)}$, and weighting function
    \begin{equation}
		w(x,u)\T:=k^N_{\hat{X}\hat{U}}(x,u)\T\left[ K^N_{\hat{X}\hat{U}}+N\lambda I_N\right]^{-1\!},\label{eq:weightfcn}
    \end{equation}
    with constant $\lambda\geq 0$, then, $\B$ satisfies CBC condition (c) of Definition~\ref{def:cbc} w.r.t. $\M$ with probability at least $1-\rho$. Note that this implies the existence of a policy $\pi\colon\X\rightarrow\U$ satisfying \eqref{eq:ineqCondForallX2}.
\end{theorem}
\begin{proof}
    We start by rewriting the inner product in \eqref{eq:ineqCondForallX} as
    \begin{equation*}
		\innerH{\B}{\mu(x,u)}{\Hilbert_{+}} =
		\innerH{\B}{\mu(x,u) - \hat\mu^N_{k_{+\!}|k_{xu}}\!(x,u)}{\Hilbert_{+}} + \innerH{\B}{\hat\mu^N_{k_{+\!}|k_{xu}}\!(x,u)}{\Hilbert_{+}}. 
    \end{equation*}
    Via \eqref{eq:empiricalInnerProd}, the latter term yields $w(x,u)\T\B(\hat{X}^+_N)$.
    For the prior term, let $\Gamma\colon(\X\times\U)\times(\X\times\U)\rightarrow\mathcal{L}(\Hilbert_{+})$ be the operator-valued positive definite kernel of $\mathcal{G}$ \cite<see, e.g.,>[Definition~1]{li2022optimal} given by $\Gamma((x,u),(x',u')):=k_{xu}((x,u),(x',u'))\,\mathrm{Id}_{\Hilbert_{+}}$, where $\mathcal{L}(\Hilbert_{+})$ is the Banach space of bounded linear operators from $\Hilbert_{+}$ to $\Hilbert_{+}$ and $\mathrm{Id}_{\Hilbert_{+}}$ is the identity operator on $\Hilbert_{+}$.
	Then, it follows from the reproducing property of $\Gamma$ \cite<cf.>[Equation~(2)ff.]{li2022optimal} that
	\begin{align*}
		\innerH{\B}{\mu(x,u)\!-\!\hat\mu^N_{k_{+\!}|k_{xu}}\!(x,u)}{\Hilbert_{+}} 
        &= \innerH{\Gamma(\cdotx,(x,u))\B}{\mu-\hat\mu^N_{k_{+\!}|k_{xu}}}{\mathcal{G}},\\
		&\leq \varepsilon \norm{\Gamma(\cdotx,(x,u))\B}_{\mathcal{G}},\\
		&= \varepsilon \sqrt{\innerH{\Gamma(\cdotx,(x,u))\B}{\Gamma(\cdotx,(x,u))\B}{\mathcal{G}}},\\
		&= \varepsilon \sqrt{\innerH{\B}{\Gamma((x,u),(x,u))\B}{\Hilbert_{+}}},\\
		&\leq \varepsilon\bar{\B}\sqrt{\norm{\Gamma((x,u),(x,u))}},\\
		&= \varepsilon\bar{\B}\sqrt{k_{xu}((x,u),(x,u))}.
	\end{align*}
	Reordering yields \eqref{eq:ineqCondForallX2}, concluding the proof.
\end{proof}
Intuitively, the conservatism introduced
due to the worst-case approach w.r.t. the unknown transition kernel $\Tr$
is captured by the term $\varepsilon\bar{\B}\kappa$, thus proportional to the radius $\varepsilon$. Ensuring that \eqref{eq:ineqCondForallX2} holds for an adequate offset $\varepsilon\bar{\B}\kappa$ gives the necessary headroom to provide guarantees for dynamics deviating from the empirical observations, captured in the CME.
As we design the CBC $B$, an upper bound $\bar{\B}$ can be estimated via analytical expressions or numerical estimation \cite{Berlinet2004RKHSProbStat,Kanagawa2018GPvsKernel}.
With the following proposition, we quantify the safety probability of a black-box system based on the results from Theorem~\ref{thm:cbc2}.
\begin{theorem}[Data-driven finite-horizon safety]\label{thm:safety}
    Consider the data-based setup in Theorem~\ref{thm:cbc2} and a safety specification $\psi_{\mathrm{safe}}=(\X_u,T)$. Suppose there exists a function $B$ satisfying the conditions in Theorem~\ref{thm:cbc2} for a constant $c\geq 0$ and policy $\pi$. If there {exists a constant} ${\eta\in[0,1)}$ such that
    \begin{itemize}
	\item[(a)] $\forall x_0\in \X_0\colon\B(x_0)\leq\eta$; and
	\item[(b)] $\forall x_u\in \X_u\colon\B(x_u)\geq{1}$; 
    \end{itemize}
    are satisfied, then, with probability at least $1-\rho$ we have
    \setlength{\belowdisplayskip}{3pt}
    $$P^\pi_{\mathrm{safe}}(\M)\geq 1- {(\eta + cT)}.$$
\end{theorem}

Theorem~\ref{thm:safety} follows trivially from Theorem~\ref{thm:cbc2} and Proposition~\ref{prop:reach}.

\section{Extension to General Classes of Temporal Logic Specifications}
\label{sec:LTL}
The results presented in the previous section can be extended to develop a data-driven approach for computing a lower bound on the satisfaction of more general specifications beyond safety.
Examples include developing AI systems satisfying liveness, fairness, and complex temporal properties in scenarios involving cooperation or human interaction.

\subsection{Linear Temporal Logic over Finite Traces}

{The work by \shortciteA{Pushpak2021CBC} introduces a model-based procedure for using CBCs to compute a lower bound on the probability of satisfying \emph{LTL specifications over finite traces} (\LTLf). The procedure comprises the following steps:
\begin{enumerate}[label=(\roman*)]
    \item The specification is negated;
    \item The negated specification is translated into a DFA;
    \item The resulting DFA is decomposed into a sequence of reachability tasks;
    \item For each individual reachability task, an upper bound on its satisfaction probability is computed using CBCs; and
    \item The individual bounds are composed, based on the structure of the DFA, to obtain a lower bound on the probability of satisfying the original \LTLf specification.
\end{enumerate}}

{This procedure relies primarily on knowledge of the specification and uses CBCs as a subroutine to establish bounds on the individual reachability probabilities.
In scenarios where a model of the system is \emph{not} available but an \LTLf specification is given, the identical procedure outlined above can be applied in conjunction with data-driven CBCs constructed using the results of Section~\ref{sec:ddcbc}.
Further technical details are omitted here, as the overall approach remains unchanged from that of \citeA{Pushpak2021CBC}, with the only difference being the substitution of model-based CBCs by their data-driven counterparts developed in Section~\ref{sec:ddcbc}.}

\subsection{Specifications Modeled via Co-B\"uchi Automata}
{In an effort to address a broader class of specifications beyond \LTLf, the work by \shortciteA{anand2024compositional} proposes a method for computing a lower bound on the probability of satisfying $\omega$-regular specifications in networked stochastic systems. However, the approach is based on a fundamental simplification: the original specification is reduced to an automaton with a co-B\"uchi acceptance condition --- that is, an acceptance criterion requiring certain states to be visited only finitely often. Consequently, the method is effectively restricted to a \emph{strict subset} of $\omega$-regular specifications.}

{The approach of \citeA{anand2024compositional} comprises the following steps:
\begin{enumerate}[label=(\roman*)]
    \item The $\omega$-regular specification is modeled as a DFA with a Rabin acceptance condition;
    \item The specification is strengthened by replacing the Rabin acceptance condition with a co-B\"uchi acceptance condition;
    \item A subset of satisfying traces with repeating cycles is extracted from the resulting automaton;
    \item These traces are decomposed into sequences of safety specifications;
    \item CBCs are employed to compute individual lower bounds on the probability of satisfying these safety specifications; and
    \item The individual bounds are combined to obtain a lower bound on the probability of satisfying the original specification.
\end{enumerate}
As for the approach proposed by \citeA{Pushpak2021CBC} reviewed previously, the above procedure relies primarily on knowledge of the specification and uses CBCs as a subroutine to obtain bounds on the individual safety probabilities.}
{Thus, the data-driven results of Section~\ref{sec:ddcbc} can be used to extend the procedure to a model-free setting. The underlying steps remain unchanged, and so the mathematical details are omitted.}

\subsection{$\boldsymbol{\omega}$-Regular Specifications}
To address the general class of $\omega$-regular specifications, we build on the concept of \emph{Streett supermartingales}, proposed by \citeA{abate2024stochastic}, and provide a data-driven version that can be established without model knowledge and based only on data from the system. 
For this, let a \emph{Streett pair}, i.e., a pair of sets $(F,I)$ with $F,I\in\borel{\X}$, be given.
The following result is an extension of Theorem~3 by \citeA{abate2024stochastic} to systems with control inputs.

\begin{proposition}[Streett supermartingale]
\label{prop:streett}
    For an MDP $\M=(\X,\X_0,\U,\Tr)$ and a Streett pair $(F,I)$, suppose there exist two functions $V\colon \X \rightarrow \mathbb{R}$, $\B\colon \X \rightarrow \mathbb{R}_{\geq 0}$, and positive constants $\chi, \nu>0$, such that for all $x\in\X$ there is a $u\in\U$ satisfying the following conditions:
\begin{itemize}
    \item[(a)] $V(x^+)\leq V(x)$ almost surely with $x^+\sim\Tr(\cdotx| X=x,\,U=u)$, if $x \in W$; 
    \item[(b)] $\E_\Tr[\, \B(X^+)\mid X=x,\,U=u\,] - \B(x) \leq -\chi$, if $x \in (F\backslash I) \cap  W$;
    \item[(c)] $\E_\Tr[\, \B(X^+)\mid X=x,\,U=u\,] - \B(x) \leq \nu$, if $x \in I \cap W$; and
    \item[(d)] $\E_\Tr[\, \B(X^+)\mid X=x,\,U=u\,] - \B(x) \leq 0$, if $x \in W \backslash(F \cup I)$,
\end{itemize}
where $W =\{x\in \X\mid V(x)\leq 1\}$.
Then, we call $\B$ a \emph{Streett supermartingale} and there exists a control policy $\pi\colon\X\rightarrow\U$ under which the MDP starting from any $x_0\in W$ almost surely either visits $F$ a finite number of times or visits $I$ infinitely often.
\end{proposition}
From the four conditions in Proposition~\ref{prop:streett}, the last three involve the computation of conditional expectations. These conditions can be replaced with their data-driven version using CMEs. The first condition, however, requires the inequality to hold almost surely. To satisfy the first condition, we raise the following assumption.

\begin{assumption}
\label{ass:streett}
    For the MDP $\M=(\X,\X_0,\U,\Tr)$, the support of $\Tr$ is known, i.e., there is a set $S(x,u)$ such that $\Tr(S(x,u)| x,u) = 1$ for all $x\in\X$ and $u\in \U$.
\end{assumption}
\begin{remark}
    When the system is directly represented with dynamics in \eqref{eq:model}, the above assumption can be satisfied by requiring that the deterministic part of the dynamics $f$ and the support $\mathbb{W}$ of $w_t$ are known, but the measure $p_w\in\mathcal{P}(\mathbb{W})$ of $w_t$ is unknown. In this case, $S(x,u) = \{f(x,u,w)\mid w\in \mathbb{W}\}$.
\end{remark}
The next theorem presents the conditions for the CME-based computation of Streett supermartingales under Assumption~\ref{ass:streett}.

\begin{theorem}
    \label{thm:streett}
    Let data $(\hat{X}_N, \hat{U}_N, \hat{X}^{+\!}_N)$ in \eqref{eq:data} from an unknown MDP $\M$ under Assumption~\ref{ass:streett} and
    kernels $k_{xu}$ and $k_{+\!}$ (characteristic) be given.
    Consider the resulting empirical CME $\hat\mu^N_{k_{+\!}|k_{xu}}$ in \eqref{eq:empiricalCME} and the ambiguity set $\ambRKHS^{N}_\varepsilon$ in \eqref{eq:ambRKHS} with confidence bound $1-\rho$.
    For a Streett pair $(F,I)$, suppose there exist two functions $V\colon \X \rightarrow \mathbb{R}$, $\B\colon \X \rightarrow \mathbb{R}_{\geq 0}$, $\B\in\Hilbert_+$, and positive constants $\chi, \nu>0$, such that for all $x\in\X$ there is a $u\in\U$ s.t.
    \begin{itemize}
    \item[(a)] $V(x^+)\leq V(x)$, for all $x^+\in S(x,u)$, if $x \in I$;
    \item[(b)] $w(x,u)\T \B(\hat{X}^+_N) - \B(x)\leq  -\chi - \varepsilon\bar{\B}\kappa(x,u)$, if $x\in (F\backslash I) \cap  W$;
    \item[(c)] $w(x,u)\T \B(\hat{X}^+_N) - \B(x)\leq \nu - \varepsilon\bar{\B}\kappa(x,u)$, if $x \in I \cap W$; and
    \item[(d)] $w(x,u)\T \B(\hat{X}^+_N) -\B(x) \leq - \varepsilon\bar{\B}\kappa(x,u)$, if $x \in W \backslash(F \cup I)$;
\end{itemize}
where $ W =\{x\in \X\mid V(x)\leq 1\}$, $\bar{\B}\geq\norm{\B}_{\Hilbert_+}$, $\kappa(x,u):=\sqrt{k_x(x,x)}\sqrt{k_u(u,u)}$, and $w(x,u)\T$ given in \eqref{eq:weightfcn}.
Note that this implies the existence of a policy $\pi\colon\X\rightarrow\U$ satisfying the conditions above.
Then, we call $\B$ a \emph{Streett supermartingale} and the MDP $\M$ starting from any $x_0\in  W$ and under the policy $\pi$ almost surely either visits $F$ a finite number of times or visits $I$ infinitely often.
\end{theorem}

Any $\omega$-regular specification can be modeled with a deterministic finite automaton with a Streett acceptance condition \cite{streett1981propositional,BK08}. The acceptance condition consists of a finite set of Streett pairs $\{(F_i,I_i), i=1,2,\ldots,p\}$. A trajectory satisfies the acceptance condition if for each $i=1,2,\ldots,p$, it either visits $F_i$ a finite number of times or visits $I_i$ infinitely often. Hence, almost-sure satisfaction of an $\omega$-regular specification can be checked by finding Streett supermartingales for each pair, where each Streett supermartingale associates a winning domain $ W_i$. Then, a lower bound on the satisfaction probability can be computed by solving reachability to the intersection of the winning domains $\cap_i  W_i$. 

\begin{theorem}
    Consider the data-driven setup in Theorem~\ref{thm:streett} and a Streett condition $\left\{\left(F_i, I_i\right): i=1, \ldots, k\right\}$. If for a shared control policy $\pi\colon\X\rightarrow\U$ each pair $(F_i,I_i)$ admits a Streett supermartingale $\B_i$ according to Theorem~\ref{thm:streett} with an associated winning domain $ W_i$, then the trajectories under $\pi$ satisfy the Streett acceptance condition almost surely from any initial state $x_0\in \cap_i  W_i$.
\end{theorem}

\section{Problem Characterization and Comparison of Solution Approaches}
\label{sec:probCharacterizationAndComparison}
For a comprehensive investigation of computational techniques, we focus on \emph{safety verification} in the rest of the paper. Therefore, the control policy is assumed to be given, and the controlled system is denoted by the same notation after eliminating the input from the mathematical expressions.   

\medskip

We start by summarizing and characterizing the optimization problem identified for computing CBCs from data, {which implements the conditions from Theorems~\ref{thm:cbc2}--\ref{thm:safety}}:
\begin{subequations}
\begin{align}
    \min_{\B, c, \eta} \quad &{\eta + cT},\label{eq:semiinf_prog_objective}\\
    \text{s.t.}\quad
    &\forall x_0\in\X_0\colon\B(x_0)\leq\eta,\label{eq:semiinf_prog_initial}\\
    &\forall x_u\in\X_u\colon\B(x_u)\geq{1},\label{eq:semiinf_prog_unsafe}\\
    &\forall x\in\X w(x)\T B(\hat{X}^+_N) -\B(x)  \leq c - \varepsilon\bar{\B}\kappa,\label{eq:semiinf_prog_kushner}\\
    & \forall x\in\X\colon \B(x)\geq 0,\label{eq:semiinf_prog_positive}\\
    &c\geq 0,\,{\eta\in[0,1)},\,\B\in\Hilbert_+,\nonumber
\end{align}\label{eq:semiinf_prog}%
\end{subequations}
with coefficients $\kappa\geq\sup_{x\in\X}\sqrt{k_x(x,x)}$\footnote{The reformulation of $\kappa$ for translation invariant kernels $k_x$ is exact.}, $\bar{\B}\geq\norm{B}_{\Hilbert_+}$, and weight function 
\begin{equation*}
    w(x)\T:=k^N_{\hat{X}}(x)\T\left[ K^N_{\hat{X}}+N\lambda I_N\right]^{-1\!}.
\end{equation*}
Clearly, \eqref{eq:semiinf_prog} constitutes an optimization problem with infinitely many constraints. Furthermore, recall that any barrier $\B\in\Hilbert_+$ can be represented based on a linear combination of spatial support vectors $\bar{x}_1,\bar{x}_2,\ldots\in\X$ and coefficients $b_i\in\R$, namely
\begin{equation}
    \B(x):=\sum_i b_i k_+(\bar x_i, x).\label{eq:representer_form}
\end{equation}
For popular choices of the kernel $k_+$ such as the Gaussian, SQExp, or Mat\'ern kernels, the semi-infinite program \eqref{eq:semiinf_prog} is generally not convex.
We elaborate on possible (partial) dualizations in Subsection~\ref{sec:monolithic}.
The optimization problem \eqref{eq:semiinf_prog} therefore belongs to the general category of \emph{non-convex semi-infinite problems} (SIPs).
There exists a collection of approaches from the robust optimization and semi-finite programming literature relevant to this problem category \shortcite{djelassi2021recent}. 
In the following subsections, we provide an overview of the approaches we tested in implementation. 
More concretely, we juxtapose the frameworks based on their theoretical complexity and report our experimental findings.
We also note that, as all the approaches discussed take \eqref{eq:semiinf_prog} as a starting point, they are all based on the fundamental Assumption~\ref{asm:BCinRKHS}.
That being said, we emphasize that this assumption can be set arbitrarily general and we do \underline{not} raise/consider any strong assumptions such as linear, affine, or polynomial dynamics that could be exploited by special solvers.

\subsection{Monolithic Solutions}\label{sec:monolithic}
Control barriers have been developed to reduce the computational burden associated with formal methods. 
In particular, they have been developed as a supposedly ``abstraction-free" alternative to address the so-called \emph{curse of dimensionality}, that arises from approaches based on the spatial partitioning of the system domain, aptly dubbed ``abstraction-based".
Over the years, a list of customary assumptions has been established in the literature to obtain feasible formulations of the barrier problem that can be efficiently solved monolithically, i.e., without spatial or spectral subdivision. 
In the model-based case, the most typical assumptions include control-affine polynomial dynamics and polynomial barriers (both of known maximum degree). In the model-free/data-driven case, additional Lipschitz bounds on the dynamics and intermediate functions \cite{Salamati2021DDCBC} are inevitable, knowledge assumed given even though the dynamics are deemed unknown.
Evidently, it is unclear how these assumptions can be rigorously justified.

\medskip

\subsubsection{Sum-of-Squares}
The \emph{sum-of-squares} (SOS) approach has gained notable popularity for establishing barriers in polynomial system models \shortcite{prajna2005sostools}, with \citeA{schon2024DRObarrier} being the first to cast \emph{data-driven} barrier problems into SOS form. This method relies on several assumptions: polynomial barrier functions, sets $\X,\X_0,\X_u\subset\R^n$ given in semi-algebraic form, and, in the data-driven case, dynamics that are polynomially dependent on the state $x$ (i.e., $\Hilbert_x$ limited to the RKHS of the polynomial kernel).
Furthermore, the computational complexity hinges directly on the maximum polynomial degree appearing in the resulting program.
There exist extensions to bases beyond polynomials, known as \emph{kernel SOS} (KSOS) \shortcite{marteau2020non}. For instance, \citeA{bagnell2015learning} restrict a kernel machine to functions of SOS form from the product RKHS $\Hilbert':=\Hilbert\otimes\Hilbert$ based on a kernel $$k'(x,x'):=\innerH{\phi(x)\otimes\phi(x)}{\phi(x')\otimes\phi(x')}{\Hilbert'}=\innerH{\phi(x)}{\phi(x')}{\Hilbert}^2.$$
We provide further examples of such methods for \emph{hard-constraining} kernel-machines in Subsection~\ref{sec:divide_and_conquer}.
Note, that in the multivariate case $n>1$ SOS functions only constitute a subset of non-negative functions, hence introducing some conservatism \cite{parrilo2003semidefinite}. Nevertheless, the SOS approach is very popular for obtaining well-behaved/convex relaxations of complex programs such as the SIP in \eqref{eq:semiinf_prog} as it avoids domain discretization and nonlinear optimization.

Although nomenclature-wise abstraction-free, the SOS approach is not immune to the curse of dimensionality, as it covertly associates a semi-definite program (SDP) for which the number of variables scales exponentially with the system dimension $n$ and the maximum polynomial degree $d$. The resulting complexity of the SOS is polynomial in the number of variables.
Additionally, solving barrier problems such as the SIP \eqref{eq:semiinf_prog} usually requires the expression of the sets $\X,\X_0,\X_u$, the constants $\eta$ {and} $c-\varepsilon\bar{\B}\kappa$, and additional Lagrange multipliers in the chosen kernel basis, which can be cumbersome (especially in an infinite basis) and increase the size of the SDP. There exist sampling-based extensions that can {(partly)} bypass this additional overhead and even exploit sparsity under the assumption of poisedness \cite{lofberg2004coefficients,cifuentes2017sampling}.
{As developing efficient SOS solvers is an active field of research, there exist many relevant recent advancements~\cite{cunis2022sequential,coey2022solving}.}
Refer to the paper by \citeA{parrilo2003semidefinite} for a more detailed dissemination of the SOS approach as well as \citeA[Sec.~V.C.1]{Pushpak2021CBC} and \citeA{schon2024DRObarrier} for its application to barrier certificates.

\medskip

\subsubsection{Dual Reformulation}
In an effort to find similar variants of the SIP \eqref{eq:semiinf_prog} that can be efficiently computed monolithically but that require less restrictive assumptions, \emph{dual reformulation} of the constraints may be considered. For example, for a polyhedral set $\X_u:=\{x_u\in\X\mid H_ux_u\leq h_u\}$ with matrix $H_u$ and vector $h_u$ of appropriate dimension, the equivalent dual to \eqref{eq:semiinf_prog_unsafe} is given by
\begin{equation}
    \min_{\lambda_u\leq0} B^\ast(H_u\T\lambda_u)
    - \lambda_u\T h_u \leq -{1}, \label{eq:dual_constr_unsafe}
\end{equation}
where $B^\ast$ is the Fenchel conjugate\footnote{The Fenchel conjugate of a function $f\colon\X\rightarrow\R$ is defined as $f^*(y)=\sup_{x\in\X} \left(y^T x-f(x)\right)$ \cite{boyd2004convex}.} of $B$. Closed-form solutions for \eqref{eq:dual_constr_unsafe} can be obtained for specific choices of the kernel $k$, as shown with the following example.
\begin{example}[Exponential kernel]\label{ex:const2_expkernel}
    Consider $\B\in\Hilbert$ with $\Hilbert$ the RKHS of the exponential kernel $k(x,x'):=\alpha e^{\beta x\T x'}$, $\alpha,\beta>0$. Then, $\B$ can be expressed in the form $\B(x)=\alpha\sum_i a_i e^{\beta x_i\T x}$, where $\{x_i\}_{i\in\mathcal{I}}\subset\X$ is an orthogonal basis of $\X$. 
    Furthermore, the Fenchel conjugate $\B^\ast$ is given by
    \begin{equation}
        \B^\ast(\lambda) = \frac{1}{\beta} \sum_i \frac{1}{\norm{x_i}}x_i\T H_u\T\lambda.\label{eq:dualconstr2_expkernel}
    \end{equation}
    For an ortho\underline{normal} basis $\{x_i\}_{i\in\mathcal{I}}$ we additionally have $\norm{x_i}=1,\forall i\in\mathcal{I}$.
    The derivation is omitted for brevity.
\end{example}
Even though this shows that we can indeed obtain convex closed-form reformulations of individual constraints for specific choices of the kernel, it is \emph{not} possible to convexify the entire optimization problem \eqref{eq:semiinf_prog} in this way. In fact, non-linear convexity of both constraints \eqref{eq:semiinf_prog_initial} \emph{and} \eqref{eq:semiinf_prog_unsafe} is contradictory.
Furthermore, the Kushner constraint \eqref{eq:semiinf_prog_kushner} is generally neither convex nor concave.
We will revisit dual reformulation in the context of local optimization problems as used in branch-and-bound optimizers in Subsection~\ref{sec:divide_and_conquer}.

\subsection{Divide-and-Conquer Solutions}\label{sec:divide_and_conquer}
The main take-away from the previous subsection is that solving the non-convex SIP \eqref{eq:semiinf_prog} in a monolithic fashion is only feasible under carefully chosen assumptions. But even then, {the solution approach suffers --- similar to abstraction-based approaches --- from the curse of dimensionality.}
For those that strive for maximal flexibility and minimal assumptions, in this subsection, we discuss approaches that (at least partially) leverage a spatial subdivision of the domain of the system/barrier. Although bearing similarities to traditional abstraction-based approaches, they leverage the barrier as a mathematical vehicle for storing information and guiding the spatial abstraction process. In comparison, abstraction-based approaches often suffer from significant memory requirements and adaptive abstraction regimes rely on ad-hoc principles such as the approximation of local Lipschitz constants \cite{esmaeil2013adaptive,kazemi2022datadriven}.

\medskip

\noindent\textbf{One-stage vs. Two-stage solutions.}
Approaches to solving the SIP \eqref{eq:semiinf_prog} can be applied in both one-stage and two-stage frameworks. One-stage methods determine valid barriers directly, while two-stage methods involve (i) \emph{candidate barrier identification} and (ii) \emph{candidate barrier verification}. For instance, the CEGIS approach is a two-stage method that iterates between learning candidate barriers and verifying them using a counterexample approach \shortcite{edwards2024fossil2}. Typically, verifying a candidate barrier is less complex than generating it, rendering two-stage methods generally more computationally efficient.
In Section~\ref{sec:algorithm}, we present an efficient sampling-based algorithm for generating barriers that are valid \emph{by design}, thus avoiding the computational challenges associated with CEGIS.
For completeness, we will discuss and compare various verification approaches in the following.

\medskip

\subsubsection{SAT/SMT Solvers}
\emph{Satisfiability} (SAT) solvers and \emph{satisfiability modulo theory} (SMT) solvers are useful algorithmic tools for determining if a given set of equations can be satisfied.
Thus, SMT solvers such as \textsc{Z3} \cite{z3} and \textsc{dReal} \shortcite{gao2013dreal} have found widespread use in applications requiring formal correctness guarantees. 
For example, \shortciteA{huang2017safety} use \textsc{Z3} to verify the safety of neural networks. 
Similarly, the recent barrier tool by \citeA{edwards2024fossil2} relies on SMT solvers to verify the soundness of generated candidate barriers.
Although the aforementioned solvers can handle (some) nonlinear expressions, their numerical complexity is prohibitive.
Whilst the exact complexity of solving an SIP such as \eqref{eq:semiinf_prog} via SMT solvers is problem dependent, the general complexity of $\delta$-SMT programs such as the ones solved by \textsc{dReal} is NP-complete \shortcite{gao2012delta}.
The involved $\delta$-decision procedures construct a symbolic representation of the problem by replacing nonlinear expressions with symbolic variables. The constraints are subsequently evaluated over the continuous domain by applying interval arithmetic. The resulting number of paths to be checked scales exponentially in the number of symbolic variables; a phenomenon known as \emph{path explosion}.

\subsubsection{Global Optimizers}
Another approach to finding a feasible solution to the SIP \eqref{eq:semiinf_prog} are global optimizers with deterministic guarantees for finding the global optimum (within tolerance) in finite time \cite{horst2013global}.
Global optimizers themselves can be subcategorized depending on their operating principle. For instance, \emph{branch-and-bound} (B\&B) optimizers such as {\textsc{BARON}} \cite{sahinidis1996baron}, 
{\textsc{SCIP}} \shortcite{bestuzheva2021scip}, and {\textsc{MAiNGO}} \shortcite{bongartz2018maingo} solve the problem by iteratively subdividing the domain and performing local convex relaxations to obtain upper and lower bounds on possible function values. 
Similar to SMT solvers, they decompose the nonlinear program by assigning symbolic variables to nonlinear expressions.
For kernel-based methods such as GPs, the number of nonlinear terms scales quadratically in the number of training samples. See \shortciteA{schweidtmann2021deterministic} for a dissemination of global optimization of GPs based on branch-and-bound algorithms. 
For the specific case of neural barrier verification, interval analysis tools such as \textsc{immrax} \shortcite{harapanahalli2024immrax} can be used.
In contrast to SMT solvers, global optimizers can provide information beyond logical feasibility and determine the actual (approximate) value of the global optimum. However, SMT solvers typically leverage combinatorial techniques such as conflict-driven clause learning (CDCL) and are generally not less well equipped to solving SIPs over continuous domains than global optimizers leveraging convex relaxations.
Notably, the tool \textsc{MAiNGO} provides dedicated relaxations for common kernel functions.
The worst-case complexity of any B\&B algorithm is exponential. However, the average complexity of the algorithm can be significantly lower on a case-by-case basis.
In our experiments using the SQExp kernel, neither the SMT solver \textsc{dReal} nor any of the previously mentioned B\&B optimizers was able to produce results in a reasonable time frame for a minimal example of the SIP \eqref{eq:semiinf_prog} involving all constraints.

\subsubsection{Compact Coverings}\label{sec:compact_coverings}
Most approaches to hard shape constraints for kernel methods use a compact covering of the domain.
For instance, the covering-based method outlined by \citeA{aubin2020hard} invokes inequality constraints on function derivatives by sampling equidistant points.
Similar to global optimizers, compact coverings leverage regularity information of the optimization problem. In fact, a compact cover abandons the iterative B\&B scheme and constitutes its limit case by forming branches according to a globally unified regularity property. Thus, compact coverings do not exclude regions of sure constraint satisfaction (that could be known from a convex relaxation) from the computations. This results in a inferior computational complexity of $\mathcal{O}(r^n)$, where $r$ is the resolution of the discretization and $n$ the number of system dimensions \cite{lasserre2001global}.
To give an intuition, for the SQExp kernel \eqref{eq:sqexp_kernel}, the resolution $r$ required to obtain an error is highly dependent on the amplitude $\sigma_f^2$ of the kernel.

\subsubsection{Other Approaches}
There are a few noteworthy approaches that do not directly fit into the aforementioned categories or are currently impractical. 
For example, the non-convex scenario optimization approach recently presented by \citeA{garatti2024non} breaks down the solution of an SIP into the identification of the least complex finite (i.e., having the fewest possible finitely many constraints) subproblem. 
However, obviously, determining this subproblem is generally difficult.
In contrast, B\&B identifies a relaxed problem via a strategical iterative procedure.

\medskip

Next, we present our approach to the SIP \eqref{eq:semiinf_prog}.
Whilst all of the previously mentioned approaches operate on the spatial domain of the SIP, choosing a kernel inducing a Fourier basis allows us to obtain a spectral representation of the problem. More specifically, we choose a finite Fourier feature expansion of the SQExp kernel that is essentially a spectral abstraction of the system.
We draw from the mathematical foundation of kernel-based methods to construct correct-by-design barriers based on an explicit mathematical characterization of the space of candidate barriers and the system's hypothesis space $\mathcal{G}$.

\section{Fourier Barrier Solution}\label{sec:algorithm}
{To address the SIP in \eqref{eq:semiinf_prog}, a central observation is made: although modeling nonlinear state evolutions, at its core, a CME corresponds to a \emph{linear} operator in RKHS. However, as characteristic RKHSs are typically infinite-dimensional, performing explicit computations within RKHS is generally not possible.
To obtain a tractable approximation, a Fourier series expansion of the characteristic squared exponential kernel is employed. This yields a finite-dimensional spectral approximation of the associated RKHS, which remains expressive enough to capture both the system dynamics and candidate barrier functions.}

\begin{figure}[htb]
    \centering
    \begin{tikzpicture}[node distance=0.4cm and 0.9cm,
            input-block/.style={rectangle, draw, fill=white, minimum height=1.0cm, align=center, rounded corners, drop shadow=gray!50},
            output-block/.style={rectangle, draw, accent, fill=accent!10, text=black, text width=4cm, minimum height=1.0cm, align=center, rounded corners, drop shadow=gray!50},
            block/.style={rectangle, draw, fill=accent!20!TACcyanwashed, text width=6.0cm, minimum height=1.0cm, align=center, rounded corners, drop shadow=gray!50},
            arrow/.style={thick,->,>=Stealth},
            inv/.style={}, 
            arrow/.style={thick,->,>=latex}
        ]

        \node[input-block] (samples) {Collect samples $\{(x^i,x_+^i)\}_{i=1}^N$};

        \node[block, right=of samples] (estimator) {Fit the estimator 
        $\E[\,f(X^+)\,|\, X=x\,\,]$
        };
        \node[block, below=of estimator, fill=accent!25!TACcyanwashed] (feature) {Construct the feature map $\phi_M\colon \X \to \mathbb{R}^{2M+1}$};
        \node[block, below=of feature, fill=accent!30!TACcyanwashed] (period_lattice) {Construct periodic lattice on $\tilde{\X}$};
        \node[block, below=of period_lattice, fill=accent!35!TACcyanwashed] (lattice) {Filter lattices on $\X, \X_0, \X_u$};
        \node[block, below=of lattice, fill=accent!40!TACcyanwashed] (parameters) {Compute coefficients $C_{\tilde{N}},
        A^{\tilde{\X}\setminus\X}_{\tilde{N}}, A^{\tilde{\X}\setminus\X_0}_{\tilde{N}}, A^{\tilde{\X}\setminus\X_u}_{\tilde{N}}$};
        \node[block, below=of parameters, fill=accent!45!TACcyanwashed] (apply-lattice) {Apply the feature map to the lattice points};
        \node[block, below=of apply-lattice, fill=accent!50!TACcyanwashed] (optimizer) {Solve the LP \eqref{eq:linear_prog}};
        \node[draw, dashed, inner sep=0.3cm, rounded corners,
        fit=(estimator)(feature)(lattice)(apply-lattice)(optimizer), label=above:{\textsc{Proposed Framework}}] {};

        \node[input-block, left=of lattice] (specification) {Specify $\X, \X_0, \X_u$};

        \node[output-block, left=of optimizer, yshift=.75cm] (output) {Barrier \mbox{certificate} $\B$};
        \node[output-block, left=of optimizer, yshift=-.75cm] (output2) {Safety probability $p_N^\pi$};

        \draw[arrow] (samples) -- (estimator);
        \draw[arrow] (specification) -- (lattice);
        \draw[arrow] (estimator) -- (feature);
        \draw[arrow] (feature) -- (period_lattice);
        \draw[arrow] (period_lattice) -- (lattice);
        \draw[arrow] (lattice) -- (parameters);
        \draw[arrow] (parameters) -- (apply-lattice);
        \draw[arrow] (apply-lattice) -- (optimizer);
        \draw[arrow] (optimizer.170) -- (output.east);
        \draw[arrow] (optimizer.190) -- (output2.east);

    \end{tikzpicture}
    \caption{Sequence of steps to generate a Fourier barrier certificate.}
    \label{fig:steps}
\end{figure}
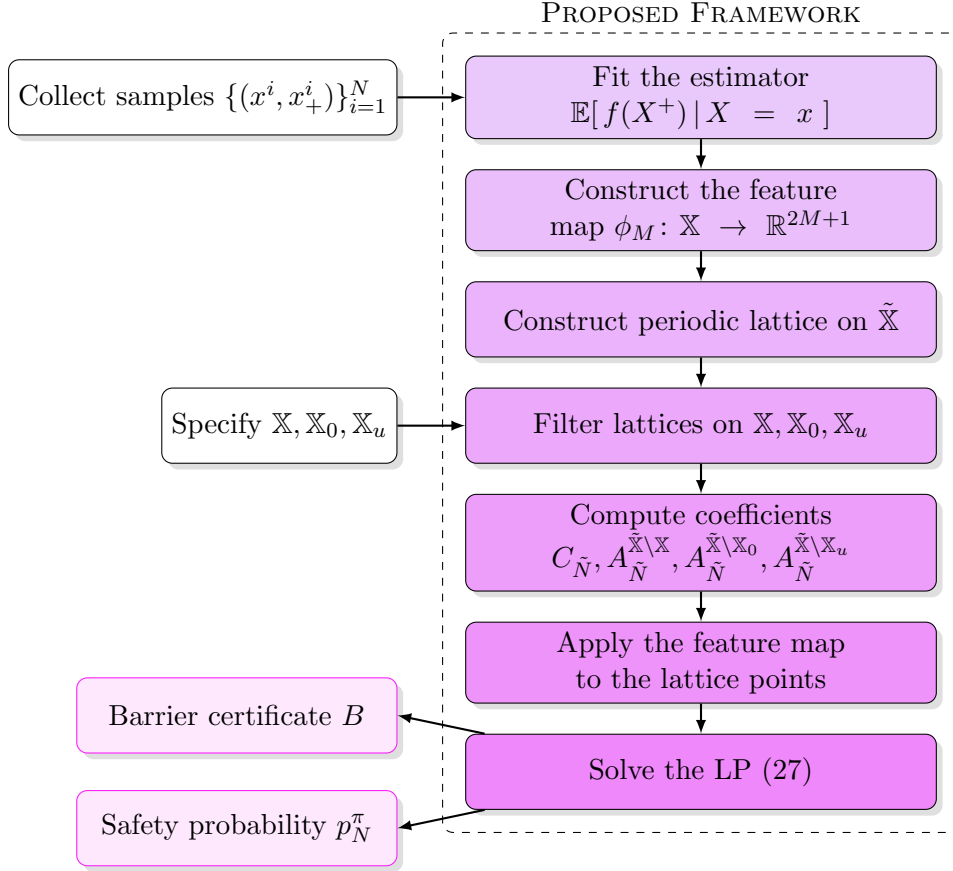

{In contrast to compact covering methods (see Subsubsection~\ref{sec:compact_coverings}), where the number of spatial support vectors scales exponentially with the dimensionality of the system, the spectral density of the squared exponential kernel is concentrated near the origin (see Figure~\ref{fig:spectral_measure_abstraction}), allowing for a compact and efficient basis requiring comparatively few spectral support vectors.
The resulting formulation is a semi-infinite linear program (SILP), which can be solved efficiently as a standard linear program (LP) using trigonometric bounding techniques~\cite{pfister2018bounding}.
Notably, the ambiguity arises here in the frequency domain, introducing uncertainty over the frequency response of the system. While this paper focuses on establishing the overall framework, quantifying the induced abstraction error and identifying principled strategies for selecting an optimal spectral basis remain open challenges for future work.}
{We provide an overview of the proposed algorithmic verification solution in Figure~\ref{fig:steps}.}

\subsection{Finite Fourier Feature Expansion}
It is well known that for popular translation-invariant kernels Bochner's theorem provides an infinite series expansion \cite{rahimi2007random}.
In the remainder of this paper, we focus on the SQExp kernel \eqref{eq:sqexp_kernel}, which admits a Fourier expansion of the form
\begin{equation}
    k(x, x') \equiv \sigma_f^2 \int_{\R^n} \mathcal{N}(d\omega\vert0,\Sigma^{{-1}}) e^{i\omega\T (P({x})-P({x'}))},\quad \Sigma:=\diag(\sigma_l)^{{2}},\label{eq:sqexp_kernel_fourier}
\end{equation}
associated with an infinite-dimensional RKHS,
where {$x\mapsto P(x)$ is an affine mapping from $\X$ onto the unit hypercube $[0,1]^n$}. {To this end, we assume $\X$ to be of hyperrectangular form.}
Recall that for the CME we have $\mu_{k_{+\!}|k_{xu}}(\Tr)(x,u)\in\Hilbert_+$ for any $(x,u)\in\X\times\U$. We deliberately restrict ourselves to designing barriers from a finite space $\Hilbert_M$, that
is spanned by a truncated Fourier basis 
\begin{equation*}
    \phi_M(x) = \sigma_f\begin{bmatrix}
        \w_0\\
        \sqrt{2}\w_1\cos(\omega_1\T P(x))\\
        \sqrt{2}\w_1\sin(\omega_1\T P(x))\\
        \vdots\\
        \sqrt{2}\w_{M}\cos(\omega_{M}\T P(x))\\
        \sqrt{2}\w_{M}\sin(\omega_{M}\T P(x))
    \end{bmatrix},
\end{equation*}
characterized by $M\in\N_{> 0}$ \emph{wavenumbers} (spatial frequencies) {$\omega_j\in\R^n$, $j=1,\ldots,M$, 
and weights $\w_0,\ldots,\w_{M}\in\R_{\geq 0}$.} 

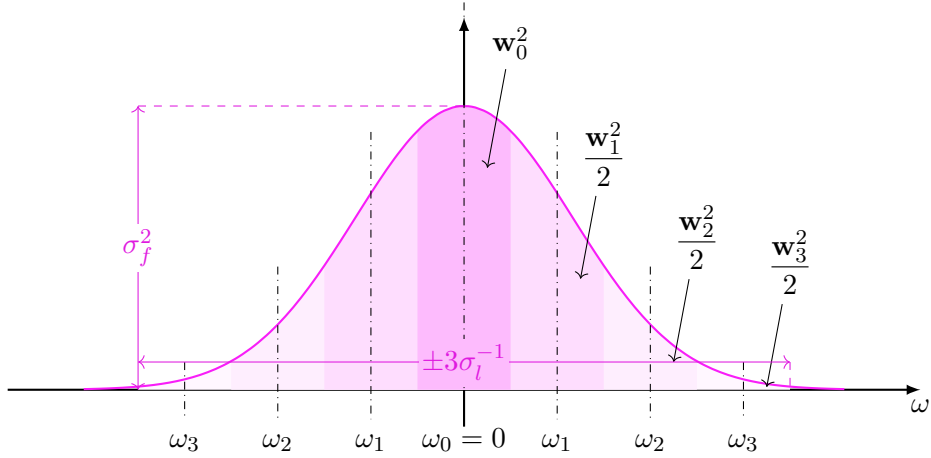
\begin{figure}
    \centering
    \begin{tikzpicture}
        \def\N{50} 
        \def\B{0};
        \def\Bs{3.0};
        \def\var{1.3}
        \def\xmax{\B+3.5*\var*\Bs};
        \def\ymin{{-0.1*gauss(\B,\B,\Bs)}};
        \def\h{0.08*gauss(\B,\B,\Bs)};

        \begin{axis}[every axis plot post/.append style={
            mark=none,domain={-(\xmax)}:{1.0*\xmax},samples=\N,smooth},
            xmin={-(\xmax)}, xmax=\xmax,
            axis/.style={>=latex},
            ymin=\ymin, ymax={1.01*gauss(\B,\B,\Bs)},
            axis lines=middle,
            axis line style=thick,
            enlarge x limits, 
            ticks=none,
            xlabel=$\omega$,
            every axis x label/.style={at={(current axis.right of origin)},anchor=north},
            width=.9\linewidth, height=0.9*\linewidth,
            y=700pt,
            clip=false,
            axis line style={-latex}
            ]

            \addplot[accent,thick,name path=B] {gauss(x,\B,\var*\Bs)};

            \path[name path=xaxis]
            (\B-\pgfkeysvalueof{/pgfplots/xmax},0) -- (\B+\pgfkeysvalueof{/pgfplots/xmax},0); 
            \addplot[accent!3.75] fill between[of=xaxis and B, soft clip={domain={\B-3*3.5/3.5*\var*\Bs}:{\B+3*3.5/3.5*\var*\Bs}}];
            \addplot[accent!7.5] fill between[of=xaxis and B, soft clip={domain={\B-3*2.5/3.5*\var*\Bs}:{\B+3*2.5/3.5*\var*\Bs}}];
            \addplot[accent!15] fill between[of=xaxis and B, soft clip={domain={\B-3*1.5/3.5*\var*\Bs}:{\B+3*1.5/3.5*\var*\Bs}}];
            \addplot[accent!30] fill between[of=xaxis and B, soft clip={domain={\B-3*.5/3.5*\var*\Bs}:{\B+3*.5/3.5*\var*\Bs}}];

            \addplot[black,dashdotted,thin]
            coordinates {({\B-3*3/3.5*\var*\Bs},{20*gauss(\B-3*3/3.5*\var*\Bs,\B,\Bs)}) ({\B-3*3/3.5*\var*\Bs},{-\h})}
            node[below=-3pt,scale=1.0] {\strut$\omega_3$};
            \addplot[black,dashdotted,thin]
            coordinates {({\B-3*2/3.5*\var*\Bs},{4*gauss(\B-3*2/3.5*\var*\Bs,\B,\Bs)}) ({\B-3*2/3.5*\var*\Bs},{-\h})}
            node[below=-3pt,scale=1.0] {\strut$\omega_2$};
            \addplot[black,dashdotted,thin]
            coordinates {({\B-3*1/3.5*\var*\Bs},{1.3*gauss(\B-3*1/3.5*\var*\Bs,\B,\Bs)}) ({\B-3*1/3.5*\var*\Bs},{-\h})}
            node[below=-3pt,scale=1.0] {\strut$\omega_1$};
            \addplot[black,dashdotted,thin]
            coordinates {(\B,{1.05*gauss(\B,\B,\Bs)}) (\B,{-\h})}
            node[below=-3pt,scale=1.0] {\strut$\omega_0=0$};
            \node[scale=1.0] (w0)
            at ({\B+.5*3/3.5*\var*\Bs},{.125*1.5}) {$\w_0^2$};
            \node (b0) at ({\B+.25*3/3.5*\var*\Bs},{.08*1.5}) {};
            \draw[->] (w0) to (b0.center);
            \addplot[black,dashdotted,thin]
            coordinates {({\B+3*1/3.5*\var*\Bs},{1.3*gauss(\B+3*1/3.5*\var*\Bs,\B,\Bs)}) ({\B+3*1/3.5*\var*\Bs},{-\h})}
            node[below=-3pt,scale=1.0] {\strut$\omega_1$};
            \node[scale=1.0] (w1)
            at ({\B+1.5*3/3.5*\var*\Bs},{.085*1.5}) {$\dfrac{\w_1^2}{2}$};
            \node (b1) at ({\B+1.25*3/3.5*\var*\Bs},{.04*1.5}) {};
            \draw[->] (w1) to (b1.center);
            \addplot[black,dashdotted,thin]
            coordinates {({\B+3*2/3.5*\var*\Bs},{4*gauss(\B+3*2/3.5*\var*\Bs,\B,\Bs)}) ({\B+3*2/3.5*\var*\Bs},{-\h})}
            node[below=-3pt,scale=1.0] at ({\B+3*2/3.5*\var*\Bs},{-\h}) {\strut$\omega_2$};
            \node[scale=1.0] (w2)
            at ({\B+2.5*3/3.5*\var*\Bs},{.055*1.5}) {$\dfrac{\w_2^2}{2}$};
            \node (b2) at ({\B+2.25*3/3.5*\var*\Bs},{.01*1.5}) {};
            \draw[->] (w2) to (b2.center);
            \addplot[black,dashdotted,thin]
            coordinates {({\B+3*3/3.5*\var*\Bs},{20*gauss(\B+3*3/3.5*\var*\Bs,\B,\Bs)}) ({\B+3*3/3.5*\var*\Bs},{-\h})}
            node[below=-3pt,scale=1.0] {\strut$\omega_3$};
            \node[scale=1.0] (w3)
            at ({\B+3.5*3/3.5*\var*\Bs},{.046*1.5}) {$\dfrac{\w_3^2}{2}$};
            \node (b3) at ({\B+3.25*3/3.5*\var*\Bs},{.001*1.5}) {};
            \draw[->] (w3) to (b3.center);

            \addplot[<->,accent!90!black]
            coordinates {({\B-3*\var*\Bs},
            .01*1.5
            ) ({\B+3*\var*\Bs},
            .01*1.5
            )};
            \addplot[dashed,thin,accent!90!black]
            coordinates {({\B+3*\var*\Bs},
            .01*1.5
            ) ({\B+3*\var*\Bs},{0})};
            \node[accent!90!black,fill=accent!30,inner xsep=1,inner ysep=2,scale=1] at (\B,
            .01*1.5
            ) {$\pm 3\sigma_l^{-1}$};
            \addplot[<->,accent!90!black]
            coordinates {({\B-3*\var*\Bs},{gauss(0,0,\var*\Bs)}) ({\B-3*\var*\Bs},{0})};
            \addplot[dashed,thin,accent!90!black]
            coordinates {({\B-3*\var*\Bs},{gauss(0,0,\var*\Bs)}) ({0},{gauss(0,0,\var*\Bs)})};
            \node[accent!90!black,fill=white,inner xsep=3,inner ysep=2,scale=1] at (\B-3*\var*\Bs,{.5*gauss(0,0,\var*\Bs)}) {$\sigma_f^2$};

        \end{axis}
    \end{tikzpicture}
    \caption{{Abstraction of a 1-dimensional Gaussian spectral measure of the SQExp kernel for $M=3$.}}
    \label{fig:spectral_measure_abstraction}
\end{figure}

{\paragraph{Frequency Selection.}
Here, we define equally spaced frequency bands on the hypercube $[-\dilation,\dilation]\subset\R^n$, with $\dilation:=6\sigma_l^{-1}/(2M+1)$, capturing 99.73\% of the kernel's spectral measure (see Figure~\ref{fig:spectral_measure_abstraction}). 
In particular, we have $\omega_j:=\diag(\zeta_j)\cdot\dilation$, with multi-indices $\zeta_j\in\N_{\geq 0}^n$.
Importantly, note that the resulting basis $\phi_M$ is generally not periodic on $\X$. Moving forward, it will prove useful to define the domain 
\begin{equation}
    \tilde\X := P^{-1}\big(\left[0, 2\pi\right]\otimes\dilation\big), \label{eq:periodic_domain}
\end{equation}
which is the smallest domain on which $\phi_M$ is periodic.
In particular, we have $\tilde\X\equiv\X$ for when $\sigma_{l,i}=6/(2\pi(2M+1))$, for all $i=1,\ldots,n$, or equivalently $\dilation=\mathbf{1}_n$, with unit vector $\mathbf{1}_n\in\R^n$.
We denote the highest frequency $f_{max}\in \N_{\geq 0}$ w.r.t. $\tilde\X$ as $f_{\text{max}}:=\max_{j=1,\ldots,M;\, i=1,\ldots,n}\zeta_{ji}$.
}

{\begin{remark}[Effect of the lengthscale]
    Note that as the lengthscale $\sigma_l$ increases --- dilating the spatial extent of the SQExp kernel $k$ in \eqref{eq:sqexp_kernel} on $\X\times\X$ --- the associated Gaussian spectral measure depicted in Figure~\ref{fig:spectral_measure_abstraction} contracts.
    This corresponds to restricting the associated RKHS to progressively smoother functions.
\end{remark}}

\begin{figure}[ht]
     \centering
     \begin{subfigure}{0.32\columnwidth}
         \centering
         \includegraphics[width=\textwidth]{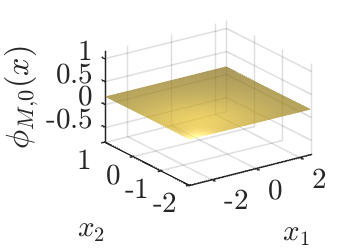}
         \caption{$\zeta_0=(0,0)$}
     \end{subfigure}
     \hfill
     \begin{subfigure}{0.32\columnwidth}
         \centering
         \includegraphics[width=\textwidth]{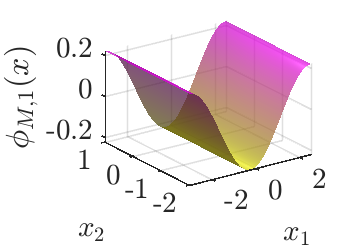}
         \caption{$\cos$, $\zeta_1=(1,0)$}
     \end{subfigure}
     \hfill
     \begin{subfigure}{0.32\columnwidth}
         \centering
         \includegraphics[width=\textwidth]{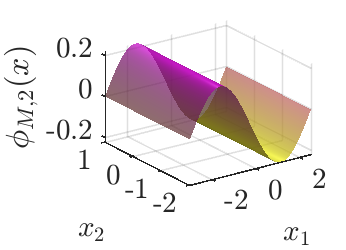}
         \caption{$\sin$, $\zeta_1=(1,0)$}
     \end{subfigure}
     \\[.5em]
     \begin{subfigure}{0.32\columnwidth}
         \centering
         \includegraphics[width=\textwidth]{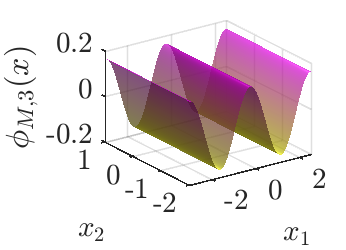}
         \caption{$\cos$, $\zeta_2=(2,0)$}
     \end{subfigure}
     \hfill
     \begin{subfigure}{0.32\columnwidth}
         \centering
         \includegraphics[width=\textwidth]{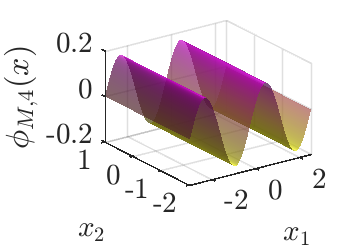}
         \caption{$\sin$, $\zeta_2=(2,0)$}
     \end{subfigure}
     \hfill
     \begin{subfigure}{0.32\columnwidth}
         \centering
         \includegraphics[width=\textwidth]{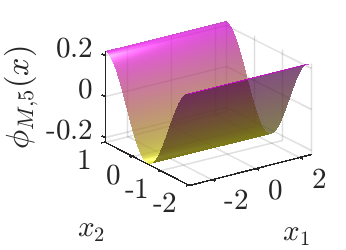}
         \caption{$\cos$, $\zeta_3=(0,1)$}
     \end{subfigure}
     \\[.5em]
     \begin{subfigure}{0.32\columnwidth}
         \centering
         \includegraphics[width=\textwidth]{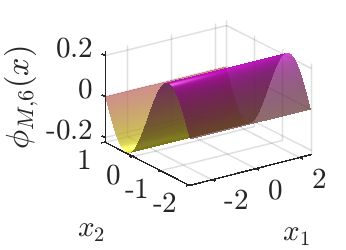}
         \caption{$\sin$, $\zeta_3=(0,1)$}
     \end{subfigure}
     \hfill
     \begin{subfigure}{0.32\columnwidth}
         \centering
         \includegraphics[width=\textwidth]{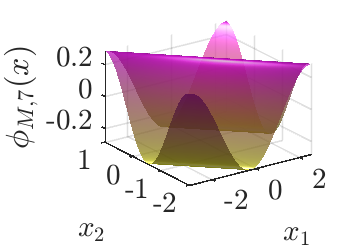}
         \caption{$\cos$, $\zeta_4=(1,1)$}
     \end{subfigure}
     \hfill
     \begin{subfigure}{0.32\columnwidth}
         \centering
         \includegraphics[width=\textwidth]{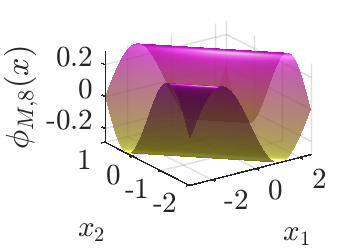}
         \caption{$\sin$, $\zeta_4=(1,1)$}
     \end{subfigure}
     \\[.5em]
     \begin{subfigure}{0.32\columnwidth}
         \centering
         \includegraphics[width=\textwidth]{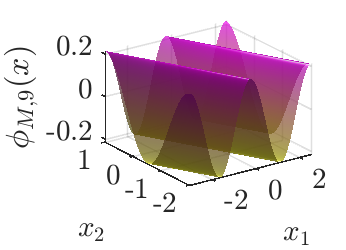}
         \caption{$\cos$, $\zeta_5=(2,1)$}
     \end{subfigure}
     \hfill
     \begin{subfigure}{0.32\columnwidth}
         \centering
         \includegraphics[width=\textwidth]{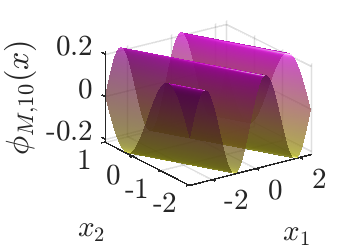}
         \caption{$\sin$, $\zeta_5=(2,1)$}
     \end{subfigure}
     \hfill
     \begin{subfigure}{0.32\columnwidth}
         \centering
         \includegraphics[width=\textwidth]{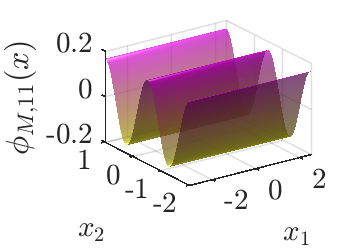}
         \caption{$\cos$, $\zeta_6=(0,2)$}
     \end{subfigure}
    \caption{Components of the spectral basis expansion used for the barrier in Section~\ref{sec:benchmark_complexspec} with a maximum frequency of $f_{\text{max}}=2$. Note that Fourier barriers are thus composed of a superposition of standing waves yielding a Fourier series expansion.}
    \label{fig:fourier_expansion}
\end{figure}

{\paragraph{Fourier CBC.}
As shown in Figure~\ref{fig:spectral_measure_abstraction}, the corresponding weights of $\phi_M$ are efficiently}
computed via the $n$-dimensional \emph{cumulative distribution function} (CDF):
\begin{equation*}
    \w_j^2 := \int_{{\omega_j-\dilation/2}}^{{\omega_j+\dilation/2}} \mathcal{N}(d\xi\vert 0,\Sigma^{{-1}}),\quad j=0,\ldots,M.
\end{equation*}
{Note that $\omega_0:=0$.}
This allows us to {expand and write the barrier $\B$ as a finite Fourier series} with $2M+1$ coefficients:
\begin{align}
    \B(x) = \phi_M(x)\T b &= \alpha_0 +\sum_{i=1}^{M} \alpha_i \cos(\omega_i\T P(x)) + \beta_i \sin(\omega_i\T P(x)),\label{eq:representer_form_spectral}\\
    \text{with}\quad b&=\begin{bmatrix}
        \frac{\alpha_0}{\sigma_f^2\w_0^2} & \frac{\alpha_1}{2\sigma_f^2\w_1^2} & \frac{\beta_1}{2\sigma_f^2\w_1^2} & \ldots & \frac{\alpha_{M}}{2\sigma_f^2\w_{M}^2} & \frac{\beta_{M}}{2\sigma_f^2\w_{M}^2}
    \end{bmatrix}\T,\nonumber
\end{align}
where the coefficients $b\in\R^{2M+1}$ act as spectral amplitudes.
Naturally, we dub barriers of the spectral form \eqref{eq:representer_form_spectral} as \emph{Fourier control barrier certificates}.
We visualize several components of an example Fourier barrier in Figure~\ref{fig:fourier_expansion}.
Compared to the typical spatial form \eqref{eq:representer_form} based on spatial support vectors $\bar{x}_1,\bar{x}_2,\ldots\in\X$, the spectral form \eqref{eq:representer_form_spectral} is a much more efficient representation for higher-dimensional systems. The number of spatial support vectors needed to construct adequately expressive barriers grows exponentially in the dimensionality of the system. We avoid this. In fact, the same principle is leveraged by memory-efficient file formats such as JPEG.

\paragraph{{Filtered Dynamics.}}
Apart from the memory efficiency of the spectral barrier form, we can apply the same idea to collapse the complex CME term in \eqref{eq:semiinf_prog_kushner} via the \emph{fast Fourier transform} (FFT). That is we obtain spectral coefficients $H\in\R^{(2M+1)\times (2M+1)}$ such that
\begin{equation}
    k^N_{\hat{X}}(x)\T\left[ K^N_{\hat{X}}+N\lambda I_N\right]^{-1\!} (\Phi_{M, \hat{X}^{+\!}}^{N})\T b \approx \phi_M(x)\T H b,\label{eq:cme_term_spectral_approx}
\end{equation}
where $\Phi_{M, \hat{X}^{+\!}}^{N} := [\phi_M(\hat{x}^+_1),\ldots,\phi_M(\hat{x}^+_N)]$.
{In fact, this approximation corresponds to applying a low-pass filter to the dynamics with a cut-off frequency of $f_{\text{max}}$.}
Whilst projection onto a truncated Fourier basis introduces an approximation error, the error decreases exponentially with the cardinality of wavenumbers in the basis \cite{rahimi2007random}. 

\begin{mdframed}[backgroundcolor=TACcyanwashed, shadow=false, hidealllines=true, leftline=true, linecolor=TACcyan, linewidth=3pt]
\vspace{-.2em}
\paragraph{{Linear SIP.}}
Substituting the barrier $B\in\Hilbert_M$ of the form \eqref{eq:representer_form_spectral} and the approximation of the CME term in \eqref{eq:cme_term_spectral_approx} into the SIP \eqref{eq:semiinf_prog}, we obtain the \emph{semi-infinite linear program} (SILP)
\begin{subequations}
\begin{align}
    \min_{b, c, \eta} \quad &{\eta + cT},\label{eq:semiinfLP_prog_objective}\\
    \text{s.t.}\quad
    &\forall x_0\in\X_0\colon\phi_M(x_0)\T b\leq\eta,\label{eq:semiinfLP_prog_initial}\\
    &\forall x_u\in\X_u\colon\phi_M(x_u)\T b\geq{1},\label{eq:semiinfLP_prog_unsafe}\\
    &\forall x\in\X\colon \phi_M(x)\T (H b - b) \leq c - \varepsilon\bar{\B}{\kappa},\label{eq:semiinfLP_prog_kushner}\\
    & \forall x\in\X\colon\phi_M(x)\T b\geq 0,\label{eq:semiinfLP_prog_positive}\\
    &c\geq 0,\,{\eta\in[0,1)},\,b\in\R^{2M+1},\nonumber
\end{align}
\label{eq:semiinfLP_prog}%
\end{subequations}
with coefficients $\kappa\geq\sigma_f$ and $\bar{\B}\geq\norm{b}_2$.
\vspace{.4em}
\end{mdframed}

\subsection{Finite-Constraint Relaxation}\label{sec:finite_constraint_relaxation}
There exist results that establish deterministic sampling-based bounds for multivariate trigonometric polynomials \cite{cifuentes2017sampling,pfister2018bounding} that can be applied to the bandlimited spectral barrier in \eqref{eq:representer_form_spectral}. This allows us to relax the SILP \eqref{eq:semiinfLP_prog} to a \emph{linear program} (LP) with finitely many constraints by sampling a finite set
${\Theta_{\hat{N}}} \subset\X$ of cardinality $\hat{N}$.
By choosing $\hat{N}= Q^n$ equidistant points that form a discrete lattice of points to cover the (state) space, the corresponding constraints can be efficiently evaluated in $\mathcal{O}(Q\log(Q))$ operations using the inverse FFT.
Note that this also substantially reduces the number of evaluations of the computationally expensive CME term in \eqref{eq:semiinf_prog_kushner}, which requires $\mathcal{O}(N^2)$ operations per evaluation. Refer to the Petersen-Middleton theorem \cite{petersen1962sampling}, which generalizes the Nyquist-Shannon sampling theorem to higher-dimensional Euclidean spaces, for further details.

We {establish} the following {lemma based on the results of} \citeA{pfister2018bounding}.

\begin{lemma}[Trigonometric bounds]\label{lem:trig_bound}
    Consider {the Fourier CBC $\B\colon\X\rightarrow\R$ in \eqref{eq:representer_form_spectral}} with maximum degree ${f_{\text{max}}}\in\N_{\geq 0}$ {and define the sampling lattice}
    \begin{equation*}
        {\Theta_{\tilde{N}}:=\left\lbrace P^{-1}\!\left(\frac{2\pi l}{\tilde{Q}}\dilation\right),\, l=0,\ldots,{\tilde{Q}}-1\right\rbrace \subset \tilde\X,\quad \tilde{N}:=\tilde{Q}^n,\quad \tilde{Q}\geq 2f_{\text{max}}+1,}
    \end{equation*}
    {with contraction/dilation coefficient $\dilation:=3\sigma_{l}^{-1}/(2M+1)\in\R^n$ and $\tilde\X$ in \eqref{eq:periodic_domain}.}
    Let $\Barrmax_{{\tilde{N}}}:=\max_{{x}\in\Theta_{{\tilde{N}}}} {\B}({x})$ and $\Barrmin_{{\tilde{N}}}:=\min_{{x}\in\Theta_{{\tilde{N}}}} {\B}({x})$. Then, we have {for all $x\in\X$,}
    \begin{equation*}
        \frac{1}{2}\left(\Barrmax_{{\tilde{N}}} + \Barrmin_{{\tilde{N}}} - C_{{\tilde{N}}}\!\left(\Barrmax_{{\tilde{N}}}-\Barrmin_{{\tilde{N}}}\right)\right)
        \leq
        {\B}({x})
        \leq
        \frac{1}{2}\left(\Barrmax_{{\tilde{N}}} + \Barrmin_{{\tilde{N}}} + C_{{\tilde{N}}}\!\left(\Barrmax_{{\tilde{N}}}-\Barrmin_{{\tilde{N}}}\right)\right)\!,
    \end{equation*}
    with a {constraint-tightening} coefficient given by
    \setlength{\belowdisplayskip}{3pt}
    \begin{equation}
        C_{{\tilde{N}}} := \left(1-\tfrac{2{f_{\text{max}}}}{{\tilde{Q}}}\right)^{-\frac{n}{2}}.\label{eq:C_coeff}
    \end{equation}
\end{lemma}
{\begin{proof}
    Note that the barrier in \eqref{eq:representer_form_spectral} is a real-valued trigonometric polynomial with maximum degree $f_{\text{max}}$ that is periodic on the domain $\tilde\X$ defined in \eqref{eq:periodic_domain}.
    According to Corollaries~1 \& 2 by \citeA{pfister2018bounding}, if $\tilde{Q}\geq 2{f_{\text{max}}}+1$ --- i.e., the minimum/Nyquist sampling rate is met --- then we have for all $x\in\tilde\X$ that
    \begin{equation*}
        \frac{1}{2}\left(\Barrmax_{\tilde{N}} + \Barrmin_{\tilde{N}} - C_{\tilde{N}}\left(\Barrmax_{\tilde{N}}-\Barrmin_{\tilde{N}}\right)\right)
        \leq
        \B(x)
        \leq
        \frac{1}{2}\left(\Barrmax_{\tilde{N}} + \Barrmin_{\tilde{N}} + C_{\tilde{N}}\left(\Barrmax_{\tilde{N}}-\Barrmin_{\tilde{N}}\right)\right).
    \end{equation*}
    As $\B$ is periodic on $\tilde\X$, the same bounds hold for any arbitrary domain $\X$.
\end{proof}}

{Via Lemma~\ref{lem:trig_bound}, we can obtain a finite-constraint relaxation of upper/lower bounds on the barrier; however, this holds only for constraints enforced on the entire periodic domain $\tilde{\X}$.
In fact, not only must \eqref{eq:semiinfLP_prog_initial} and \eqref{eq:semiinfLP_prog_unsafe} be selectively enforced only on $\X_0$ and $\X_u$, respectively, but the state space $\X$ may differ from $\tilde{\X}$ too, affecting the relaxation of \eqref{eq:semiinfLP_prog_kushner} and \eqref{eq:semiinfLP_prog_positive}.
}

\paragraph{Local Relaxation.}
Without loss of generality, let $\S\in\{\X_0,\X_u,\X\}$.
According to \citeA[Lemma~2]{pfister2018bounding} we have for $\tilde{Q}\geq 2f_{\text{max}}+1$ that for all $x\in\tilde{\X}$
\begin{equation*}
    \B(x) = \frac{1}{\tilde{N}} \sum_{\bar{x}\in\Theta_{\tilde{N}}} \B(\bar{x}) D^n_{f_{\text{max}},\tilde{Q}-f_{\text{max}}}(x-\bar{x}),
\end{equation*}
with $D^n_{a,b}\colon\R^n\to\R$ the Vallée-Poussin kernel,
defined as
\begin{equation*}
    D^n_{a,b}(z) := \frac{1}{(b-a)^n} \prod_{i=1}^n \frac{\sin(\frac{b+a}{2}z_i)\sin(\frac{b-a}{2}z_i)}{\sin^2(\frac{z_i}{2})}.
\end{equation*}
For the set $\S$, we have for all $x\in\tilde{\X}$
\begin{equation}
    \B(x) \leq \Barrmax_{\tilde{N}}^{\S} \underbrace{\frac{1}{\tilde{N}} \!\sum_{\bar{x}\in\Theta_{\tilde{N}}} 
    \!\!D^n_{f_{\text{max}},\tilde{Q}-f_{\text{max}}}\!(x-\bar{x})}_{(a)}
    + \left(\!\Barrmax_{\tilde{N}}^{\tilde{\X}\setminus\S} \!-\! \Barrmax_{\tilde{N}}^{\S}\!\right)\! \underbrace{\frac{1}{\tilde{N}} \!\sum_{\bar{x}\in\Theta_{\tilde{N}}\setminus\S} 
    \!\!D^n_{f_{\text{max}},\tilde{Q}-f_{\text{max}}}\!(x-\bar{x})}_{(b)},\label{eq:separation}
\end{equation}
where $\Barrmax_{\tilde{N}}^{\S} :=\max_{x\in \Theta_{\tilde{N}}\cap\S} \left\lbrace\phi_M(x)\T b\right\rbrace$ and $\Barrmax_{\tilde{N}}^{\tilde{\X}\setminus\S} :=\max_{x\in \Theta_{\tilde{N}}\setminus\S} \left\lbrace\phi_M(x)\T b\right\rbrace$ are computed on the lattice.
We can bound (a) as in \citeA[Lemma~4]{pfister2018bounding}:  $$\frac{1}{\tilde{N}} \sum_{\bar{x}\in\Theta_{\tilde{N}}} D^n_{f_{\text{max}},\tilde{Q}-f_{\text{max}}}(x-\bar{x})\leq C_{\tilde{N}},$$
where $C_{\tilde{N}}$ is the constraint-tightening coefficient from Lemma~\ref{lem:trig_bound}.
Term (b) can be computed numerically using, e.g., particle swarm optimization; we write
\begin{equation}
    A^{\tilde{\X}\setminus\S}_{\tilde{N}}:=\frac{1}{\tilde{N}} \sum_{\bar{x}\in\Theta_{\tilde{N}}\setminus\S} D^n_{f_{\text{max}},\tilde{Q}-f_{\text{max}}}(x-\bar{x}).\label{eq:A_coeff}
\end{equation}

Based on this decomposition, we establish upper and lower bounds akin Lemma~\ref{lem:trig_bound} for subsets $\S\subset\tilde{\X}$. The following result is based on \citeA[Corollaries~1~\&~2]{pfister2018bounding}, and the full proof can be found in Appendix~\ref{app:proof_lem_trig_bound_local}.
\begin{lemma}[{Local trigonometric bounds}]\label{lem:trig_bound_local}

    Consider the Fourier CBC $\B\colon\X\rightarrow\R$ in \eqref{eq:representer_form_spectral} with maximum degree $f_{\text{max}}\in\N_{\geq 0}$ and the sampling lattice $\Theta_{\tilde{N}}$ in Lemma~\ref{lem:trig_bound} on $\tilde\X$.
    For a subset $\S\subset\tilde{\X}$, let $\Barrmax_{\tilde{N}}^\S:=\max_{x\in\Theta_{\tilde{N}}\cap\S} \B(x)$, $\Barrmin_{\tilde{N}}^\S:=\min_{x\in\Theta_{\tilde{N}}\cap\S} \B(x)$, $\Barrmax_{\tilde{N}}^{\tilde{\X}\setminus\S}:=\max_{x\in\Theta_{\tilde{N}}\setminus\S} \B(x)$, and $\Barrmin_{\tilde{N}}^{\tilde{\X}\setminus\S}:=\min_{x\in\Theta_{\tilde{N}}\setminus\S} \B(x)$. Then, we have for all $x\in\S$,
    \begin{align*}
        &\frac{1}{2}\left(\Barrmax_{\tilde{N}}^\S + \Barrmin_{\tilde{N}}^\S - C_{{\tilde{N}}}\!\left(\Barrmax_{{\tilde{N}}}^\S-\Barrmin_{{\tilde{N}}}^\S\right)\right)+ A^{\tilde{\X}\setminus\S}_{\tilde{N}}\left(\Barrmin_{\tilde{N}}^{\tilde{\X}\setminus\S}-\Barrmin_{\tilde{N}}^{\S}\right)
        \leq
        \B(x)\\
        &\hspace{100pt}\leq \frac{1}{2}\left(\Barrmax_{\tilde{N}}^\S + \Barrmin_{\tilde{N}}^\S + C_{{\tilde{N}}}\!\left(\Barrmax_{{\tilde{N}}}^\S-\Barrmin_{{\tilde{N}}}^\S\right)\right)+ A^{\tilde{\X}\setminus\S}_{\tilde{N}}\left(\Barrmax_{\tilde{N}}^{\tilde{\X}\setminus\S}-\Barrmax_{\tilde{N}}^{\S}\right)\!,
    \end{align*}
    with the constraint-tightening coefficients $C_{\tilde{N}}$ and $A^{\tilde{\X}\setminus\S}_{\tilde{N}}$ defined in \eqref{eq:C_coeff} and \eqref{eq:A_coeff}, respectively.
\end{lemma}

Via Lemma~\ref{lem:trig_bound_local}, we obtain a finite-constraint relaxation of the {SILP \eqref{eq:semiinfLP_prog}} based on the following constraint-tightening reasoning ($\S\in\{\X_0,\X_u,\X\}$):
\begin{equation*}
    \frac{1}{2}\big(\Barrmax_{{\tilde{N}}}^{{\S}} + \Barrmin_{{\tilde{N}}}^{{\S}} - C_{{\tilde{N}}}\left(\Barrmax_{{\tilde{N}}}^{{\S}}-\Barrmin_{{\tilde{N}}}^{{\S}}\right)\big) + A^{\tilde{\X}\setminus\S}_{\tilde{N}}\left(\Barrmin_{\tilde{N}}^{\tilde{\X}\setminus\S}-\Barrmin_{\tilde{N}}^{\S}\right) \geq \mathrm{lb}
    \quad\Longrightarrow\quad
    \forall x\in\S\colon {\B}(x) \geq \mathrm{lb},
\end{equation*}
for some desired lower bound $\mathrm{lb}\in\R$.
Thus, we obtain strengthened constraints of the form 
\begin{equation*}
    \Barrmin_{{\tilde{N}}}^{{\tilde{\X}}} \geq
    \frac{2\mathrm{lb} + (C_{{\tilde{N}}}-1)\Barrmax_{{\tilde{N}}}^{{\S}} - 2A^{\tilde{\X}\setminus\S}_{\tilde{N}}\Barrmin_{\tilde{N}}^{\tilde{\X}\setminus\S}}{C_{{\tilde{N}}}-2A^{\tilde{\X}\setminus\S}_{\tilde{N}}+1}.
\end{equation*}
For an upper bound $\forall x\in\S\colon {\B}(x) \leq \mathrm{ub}$, we obtain analogous constraints of the form
\begin{equation*}
    \Barrmax_{{\tilde{N}}}^{{\S}} \leq 
    \frac{2\mathrm{ub} + (C_{{\tilde{N}}}-1)\Barrmin_{{\tilde{N}}}^{{\S}} - 2A^{\tilde{\X}\setminus\S}_{\tilde{N}}\Barrmax_{\tilde{N}}^{\tilde{\X}\setminus\S}}{C_{{\tilde{N}}}-2A^{\tilde{\X}\setminus\S}_{\tilde{N}}+1}.
\end{equation*}

\begin{remark}\label{rem:set_scaling}
    Note that the separation of $\S$ and $\tilde{\X}\setminus\S$ in \eqref{eq:separation} is arbitrary. Instead, we can also choose, e.g., a superset of $\S$ for (a), shrinking $(b)$ and yielding a potentially less conservative bound, as most of the functional mass of the Vallée-Poussin kernel is assigned locally around $\bar x$.
    In practice, we compute the terms for a superset of $\S$ inflated by, e.g., 2\%, yielding an only marginal contribution of \eqref{eq:separation}(b) to the local bound on the barrier $\B$.
\end{remark}


\subsection{Linear Program}
We are ready to present the LP relaxation of the SILP \eqref{eq:semiinfLP_prog}.
Additionally to {$\smash{\Theta_{\hat{N}}}$ and ${\smash{\Theta_{\tilde{N}}}}$} we form discrete sets $\{ x_0^{(1)}, \ldots, x_0^{(\hat{N}_0)} \} \subset \X_0$ and $\{ x_u^{(1)}, \ldots, x_u^{(\hat{N}_u)} \} \subset \X_u$ of cardinality $\hat{N}_0,\hat{N}_u\in\N$.
For given $\bar{\B}$, {$C_{\tilde{N}}, A^{\tilde{\X}\setminus\X_0}_{\tilde{N}}, A^{\tilde{\X}\setminus\X_u}_{\tilde{N}}$, and $A^{\tilde{\X}\setminus\X}_{\tilde{N}}$},
we obtain the following LP: 
\begin{bluebox}
    \begin{equation}
        \begin{alignedat}{3}
                  & \min_{\substack{b, c, \eta\\{
                  \Barrmin_{{\tilde{N}}}^{\X_0},\Barrmax_{{\tilde{N}}}^{\X_u},\Barrmin_\Delta^{{\X}},\Barrmax_{{\tilde{N}}}^{{{\X}}}}\\
                  {\Barrmax_{\tilde{N}}^{\tilde{\X}\setminus\X_0},\Barrmin_{\tilde{N}}^{\tilde{\X}\setminus\X_u},\Barrmin_{{\tilde{N}}}^{{\tilde{\X}\setminus\X}},\Barrmax_\Delta^{\tilde{\X}\setminus\X}}
                  }} \quad &                                            & \eta + cT,                                          &                       &
            \\
                  & \text{subject to}\quad
                  &                                                                                                                                & {\Barrmin_{{\tilde{N}}}^{\X_0}\leq}\phi_M(x_0^{(i)})\T b\leq\hat{\eta}, \quad &                                                     & i=1,\ldots,\hat{N}_0,
            \\
                  &                                                                                                                                &                                            & \hat{\gamma}\leq\phi_M(x_u^{(i)})\T b{\leq\Barrmax_{{\tilde{N}}}^{\X_u}},
            \quad &                                                                                                                                & i=1,\ldots,\hat{N}_u,
            \\
                  &                                                                                                                                &                                            & {\Barrmin_\Delta^{{\X}}\leq}\phi_M(x^{(i)})\T(Hb - b) \leq \hat{\Delta},
            \quad &                                                                                                                                & i=1,\ldots,\hat{N},
            \\
                  &                                                                                                                                &                                            & \hat{\xi}\leq\phi_M(x^{(i)})\T b{\leq\Barrmax_{{\tilde{N}}}^{{{\X}}}},
            \quad &                                                                                                                                & i=1,\ldots,\hat{N},
            \\
                  &                                                                                                                                &                                            & {\phi_M(x^{(i)})\T b\leq\Barrmax_{\tilde{N}}^{\tilde{\X}\setminus\X_0},}
            \quad &                                                                                                                                & {i=1,\ldots,\tilde{N}-\hat{N}_0,}                                                                                                      \\
                  &                                                                                                                                &                                            & {\Barrmin_{\tilde{N}}^{\tilde{\X}\setminus\X_u}\leq\phi_M(x^{(i)})\T b,}
            \quad &                                                                                                                                & {i=1,\ldots,\tilde{N}-\hat{N}_u,}                                                                                                      \\
            &                                                                                                                                &                                            & {\Barrmin_{{\tilde{N}}}^{{\tilde{\X}\setminus\X}}\leq\phi_M(x^{(i)})\T b,}
            \quad &                                                                                                                                & {i=1,\ldots,{\tilde{N}}-\hat{N},}                                                                                                        \\
                &                                                                                                                                &                                            & {\phi_M(x^{(i)})\T (Hb - b)\leq\Barrmax_\Delta^{\tilde{\X}\setminus\X},}
            \quad &                                                                                                                                & {i=1,\ldots,{\tilde{N}}-\hat{N},}                                                                                                        \\
                  &                                                                                                                                &                                            & c\geq 0,\,{\eta\in[0,1)},\, b\in\R^{2M+1},       &                       &
            \label{eq:linear_prog}%
        \end{alignedat}
    \end{equation}
    \setlength{\abovedisplayskip}{10pt}
    with $\kappa\geq\sigma_f$, $\bar{\B}\geq\norm{b}_2$, and constraint-tightening coefficients
    \begin{align*}
        \hat{\eta}   & := {\tfrac{2\eta + (C_{{\tilde{N}}}-1)\Barrmin_{{\tilde{N}}}^{\X_0}{-2A^{\tilde{\X}\setminus\X_0}_{\tilde{N}} \Barrmax_{\tilde{N}}^{\tilde{\X}\setminus\X_0}}}{C_{{\tilde{N}}}{-2A^{\tilde{\X}\setminus\X_0}_{\tilde{N}}}+1}},
        &&\hat{\gamma} := {\tfrac{2 + (C_{{\tilde{N}}}-1)\Barrmax_{{\tilde{N}}}^{\X_u}{-2A^{\tilde{\X}\setminus\X_u}_{\tilde{N}} \Barrmin_{\tilde{N}}^{\tilde{\X}\setminus\X_u}}}{C_{{\tilde{N}}}{-2A^{\tilde{\X}\setminus\X_u}_{\tilde{N}}}+1}},\\
        \hat{\Delta} & := {\tfrac{2(c - \varepsilon\bar{\B}\kappa) + (C_{{\tilde{N}}}-1)\Barrmin_\Delta^\X-2A^{\tilde{\X}\setminus\X}_{\tilde{N}}\Barrmax_{\Delta}^{\tilde{\X}\setminus\X}}{C_{{\tilde{N}}}{-2A^{\tilde{\X}\setminus\X}_{\tilde{N}}}+1}},
        &&\hat{\xi} := {\tfrac{(C_{{\tilde{N}}}-1)\Barrmax_{{\tilde{N}}}^{{{\X}}}{-2A^{\tilde{\X}\setminus\X}_{\tilde{N}}} \Barrmin_{\tilde{N}}^{\tilde{\X}\setminus\X}}{C_{{\tilde{N}}}{-2A^{\tilde{\X}\setminus\X}_{\tilde{N}}}+1} }.
\end{align*}
\end{bluebox}
Any solution to the LP \eqref{eq:linear_prog} that satisfies $\norm{b}_2\leq\bar{\B}$ is a feasible solution to the SIP~\eqref{eq:semiinf_prog}.
We provide more details on the derivation of the LP \eqref{eq:linear_prog} from the SILP \eqref{eq:semiinfLP_prog} via the constraint-tightening procedure outlined in Subsection~\ref{sec:finite_constraint_relaxation} in Appendix~\ref{app:silp_to_lp_derivation}.

{
\begin{remark}[Computational complexity]
    The LP in \eqref{eq:linear_prog} scales most critically with the periodic lattice size $\tilde{N}$ and the chosen Fourier barrier basis size $M$ (see Figure~\ref{fig:Barr3_ablation_study_runtime} for an ablation study). 
    The computational complexity of LPs differs between solution algorithms:
    While the simplex algorithm admits an exponential worst-case complexity, its practical performance is typically much better \cite{spielman2004smoothed}.
    Interior-point methods, on the other hand, have polynomial worst-case complexity in the problem size.

    The matrix $H$ in \eqref{eq:cme_term_spectral_approx} is computed via an FFT–based decomposition.
    The FFT has complexity $\bigO(Q \log Q)$, where $Q$ denotes the lattice resolution, which depends on the maximum frequency $f_{\max}$ considered.
    In our setting, $Q \geq (2 f_{\max}+1)^2$ by the Nyquist–Shannon sampling theorem.
    Since $H$ is computed only once during LP construction, this step contributes only a minor fraction of the overall runtime compared to solving the LP, which dominates the overall runtime in our experiments.
\end{remark}}

\section{Experimental Results}\label{sec:numerical_studies}
We demonstrate the proposed Fourier barrier approach on {two} benchmarks.
{To this end, we developed the toolbox \lucid \cite{lucid}, which implements the framework and theory introduced in this paper and is used to carry out all reported experiments.}
For all examples, we draw $N=1{\small,}000$ samples from the unknown system. 
The inverse of the Gram matrix is computed with $\lambda = 10^{-5}$.
To construct the finite-constraint LP \eqref{eq:linear_prog}, we generate a discrete lattice of $\hat{N}$ points {on $\X$} (as reported in the individual benchmarks). For simplicity, we use a rectangular sampling lattice, leaving the exploration of more complex (optimal) sampling lattices for future work.
The numerical results reported are obtained on a machine running an AMD Ryzen 9 5950X 16-core CPU and 64 GB of memory. 
{\lucid uses} a serial implementation to generate the LP. For solving the LP, we use \textsc{Gurobi} \cite{gurobi}, which is allowed to execute solution methods concurrently by default.
We set its feasibility tolerance to the minimum of $10^{-8}$, keeping the remaining settings at default.

\subsection{Complex Safety Specification}\label{sec:benchmark_complexspec}
\begin{figure}[ht]
     \centering
     \begin{subfigure}{0.49\columnwidth}
         \centering
         \includegraphics[width=\textwidth]{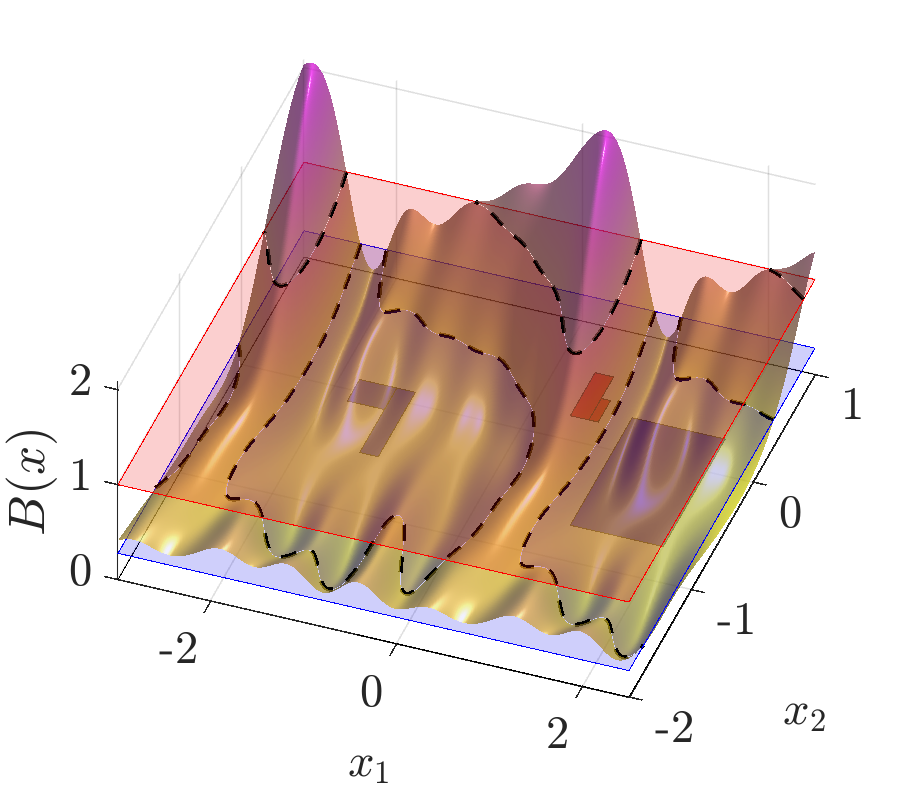}
         \caption{$M=35$, ${p_N=36.3}\%$}
     \end{subfigure}
     \hfill
     \begin{subfigure}{0.49\columnwidth}
         \centering
         \includegraphics[width=\textwidth]{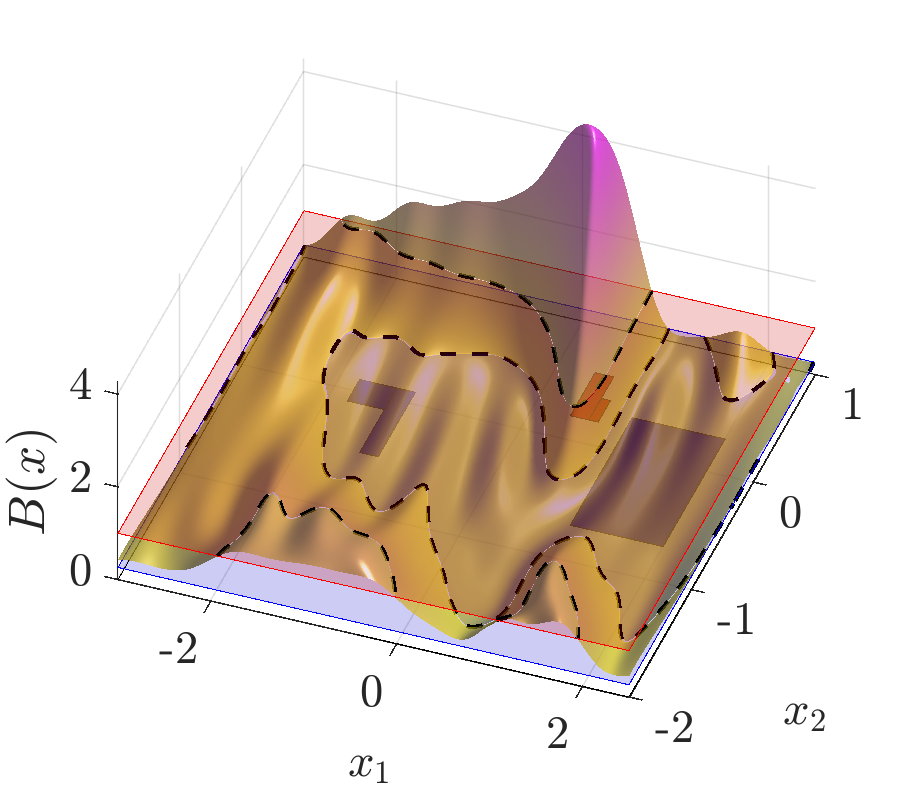}
         \caption{$M={99}$, ${p_N=50.0}\%$}
         \label{fig:probOverRadius1}
     \end{subfigure}
    \caption{Robustified CBCs and level sets {at} ${\B(x)=1}$ (in red) and $\eta$ (in blue) computed for the benchmark in Section~\ref{sec:benchmark_complexspec}, robustified with $\varepsilon\bar{\B}\sigma_f={0.007}$ and with
        (a) $M=35$ wavenumbers or
        (b) $M={99}$ wavenumbers.
        }
    \label{fig:CSBarr3}
\end{figure}

The first benchmark is inspired by the example \texttt{Barr\textsubscript{3}} from the deterministic benchmarks by \shortciteA{abate2021fossil}, which we extend by adding stochastic noise $w_t\sim\mathcal{N}(\cdotx\vert 0,0.01I_2)$ to arrive at the two-dimensional nonlinear stochastic dynamics
\begin{equation*}
    \begin{bmatrix}
        {x}_{1, t+1}\\
        {x}_{2, t+1}
    \end{bmatrix}
    = {\begin{bmatrix}
        {x}_{1, t}\\
        {x}_{2, t}
    \end{bmatrix} + \tau} \begin{bmatrix}
        {x}_{2, t}\\
        \frac{1}{3} {x}^3_{1, t} - {x}_{1,t} - {x}_{2,t} 
    \end{bmatrix} + w_t,
\end{equation*}
{with $\tau:=0.1$.}
We illustrate the setup in Figure~\ref{fig:CSBarr3}.
The goal is to compute the probability that the system initialized in $[{x}_{1,0},{x}_{2,0}]\T\in \X_0$ (blue regions) does not enter the unsafe regions $\X_u$ (in red) within $T={5}$ time steps. Note that the specified regions $\X_0$ and $\X_u$ are in close proximity to each other, rendering the barrier synthesis particularly challenging.

\paragraph{{Setup.}}
We generate the LP based on a fixed lattice {${\smash{\Theta_{\tilde{N}}}}$ with a resolution of $\tilde{Q}= 40\cdot(2f_{\text{max}}+1)$ (i.e., 40-times Nyquist sampling rate), leading to a periodic lattice of size $\tilde{N}=\tilde{Q}^2$.}
To demonstrate the expressiveness of different spectral bases, we synthesize barriers for (a) $M=6^2-1$ and (b) $M={10}^2-1$ wavenumbers {(with $f_{\mathrm{max}}=6$ and $15$, respectively),} resulting in barriers with $2M+1=71$ and ${199}$ coefficients $b_0,b_1,\ldots\in\R$, respectively.
We fix the kernel hyperparameters to $\sigma_f={1}$ and $\sigma_l={[2.993,4.629]}$ {for the input kernel $k_x$, and obtain $\sigma_{l}=[0.143, 0.358]$ (resp. $\sigma_{l}=[0.326, 0.327]$) for the output kernel $k_+$ via hyperparameter optimization. 
{For the computation of the constraint-tightening coefficients, we inflate the sets by 2\% (cf. Remark~\ref{rem:set_scaling}).}
We further use} an upper bound $\bar B={7}$ and $\varepsilon=0.001$ for both experiments.

\begin{figure}
    \centering
    \begin{subfigure}{0.49\columnwidth}
        \centering
        \includegraphics[width=\textwidth]{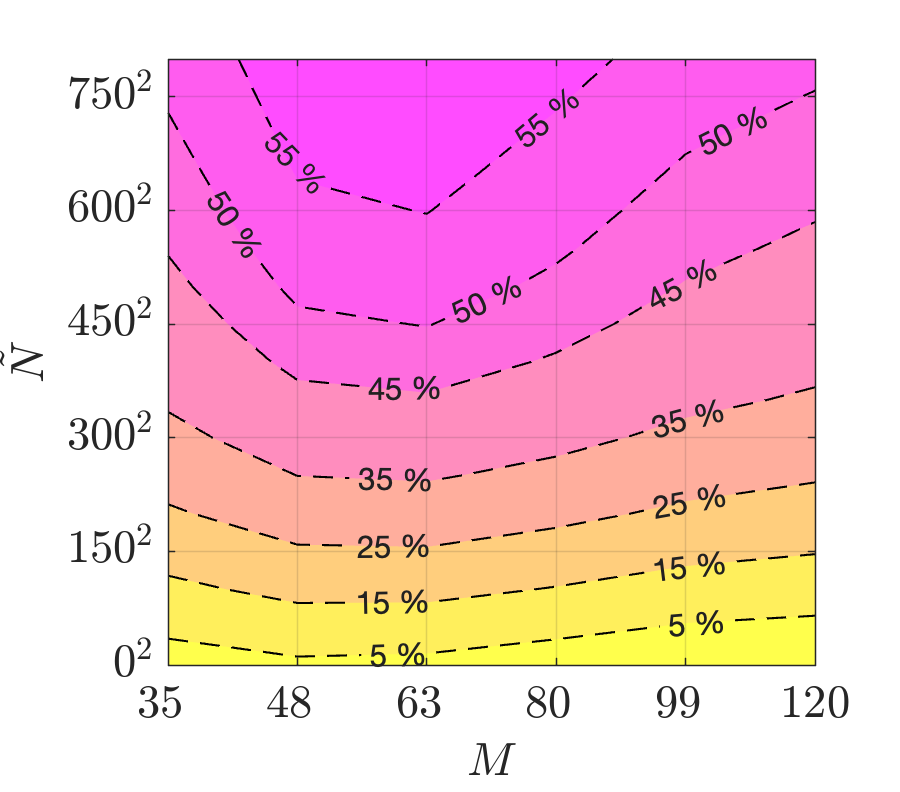}
        \caption{Safety probability $p_N$}
    \end{subfigure}
    \begin{subfigure}{0.49\columnwidth}
        \centering
        \includegraphics[width=\textwidth]{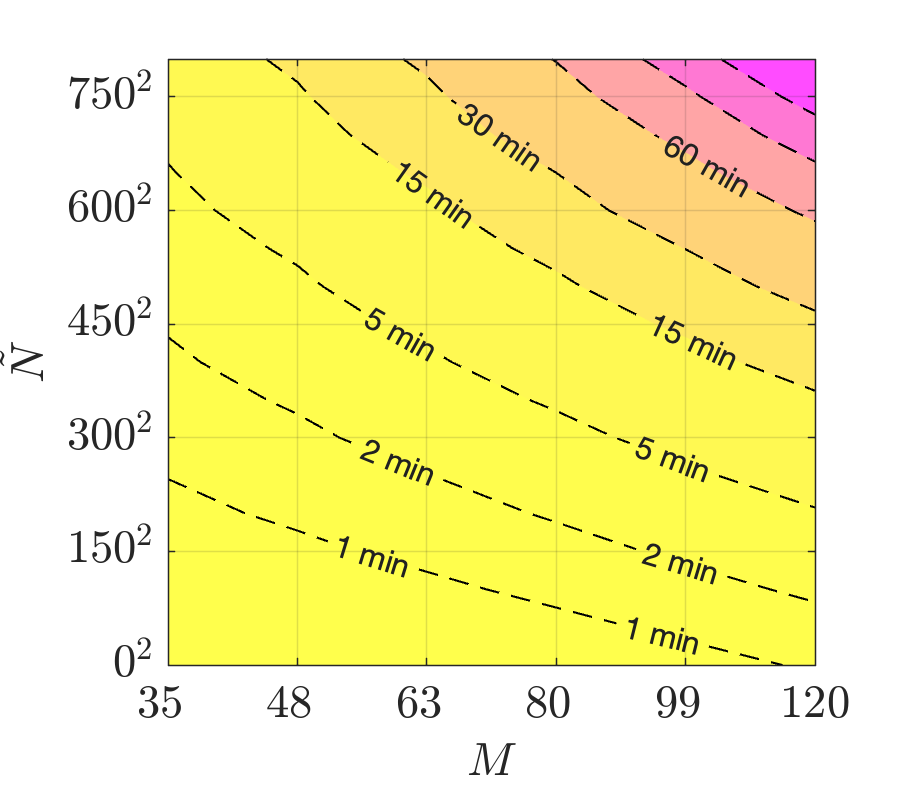}
        \caption{Runtime in minutes}
        \label{fig:Barr3_ablation_study_runtime}
    \end{subfigure}
    \caption{{Ablation study on the influence of $M$ and $\tilde N$ for the benchmark in Section~\ref{sec:benchmark_complexspec}.}}
    \label{fig:Barr3_ablation_study}
\end{figure}

\paragraph{{Results.}}
We solve the corresponding LPs and obtain the barriers shown in Figure~\ref{fig:CSBarr3}.
For (a), we find $\eta={0.277}$ and $c={0.072}$, {meaning that the identified barrier certifies a lower bound on the safety probability of the system of at least $36.3\%$ (via the formula in Theorem~\ref{thm:safety}).} 
The barrier basis (b) is richer, enabling us to find $\eta={0.263}$ and $c={0.047}$,
leading to a robust lower bound on the safety probability of the system of at least ${50.0}\%$. 
For comparison, we estimate the true safety probability ${P_{\mathrm{safe}}}$ using Monte Carlo simulation. We run $10{\small,}000$ simulations {starting from a fixed initial state $[{x}_{1,0},{x}_{2,0}]=[1,0.5]$ with a maximum length of $T$ time steps.}
Using Chebychev's inequality with a confidence level of 90\% we obtain
an estimated satisfaction probability of {$85-88\%$}.
Note that even for an infinitely complex barrier (i.e., $M\rightarrow\infty$) and data $N\rightarrow\infty$, this latent probability might not be attainable due to the inherent conservatism of the general barrier approach. 
This is in contrast to abstraction-based approaches, for which the lower bound converges to the true solution as the {partitioning of the state space is refined and more data is accumulated \cite{laurenti2023unifying}.
The effect is particularly strong here due to the closeness of the safe and unsafe regions (see Figure~\ref{fig:CSBarr3}).
See Figure~\ref{fig:Barr3_ablation_study} for an ablation study on the influence of $M$ and $\tilde N$ on the certified safety probability $p_N$ and runtime ($\varepsilon=0$).}

\subsection{Overtaking Scenario Featuring a Neural Network Controller}\label{sec:benchmark_overtaking}
\begin{figure}
    \centering
    \def\svgwidth{\linewidth}
\begingroup%
  \makeatletter%
  \providecommand\color[2][]{%
    \errmessage{(Inkscape) Color is used for the text in Inkscape, but the package 'color.sty' is not loaded}%
    \renewcommand\color[2][]{}%
  }%
  \providecommand\transparent[1]{%
    \errmessage{(Inkscape) Transparency is used (non-zero) for the text in Inkscape, but the package 'transparent.sty' is not loaded}%
    \renewcommand\transparent[1]{}%
  }%
  \providecommand\rotatebox[2]{#2}%
  \newcommand*\fsize{\dimexpr\f@size pt\relax}%
  \newcommand*\lineheight[1]{\fontsize{\fsize}{#1\fsize}\selectfont}%
  \ifx\svgwidth\undefined%
    \setlength{\unitlength}{1310.00006104bp}%
    \ifx\svgscale\undefined%
      \relax%
    \else%
      \setlength{\unitlength}{\unitlength * \real{\svgscale}}%
    \fi%
  \else%
    \setlength{\unitlength}{\svgwidth}%
  \fi%
  \global\let\svgwidth\undefined%
  \global\let\svgscale\undefined%
  \makeatother%
  \begin{picture}(1,0.48396943)%
    \lineheight{1}%
    \setlength\tabcolsep{0pt}%
    \put(0,0){\includegraphics[width=\unitlength,page=1]{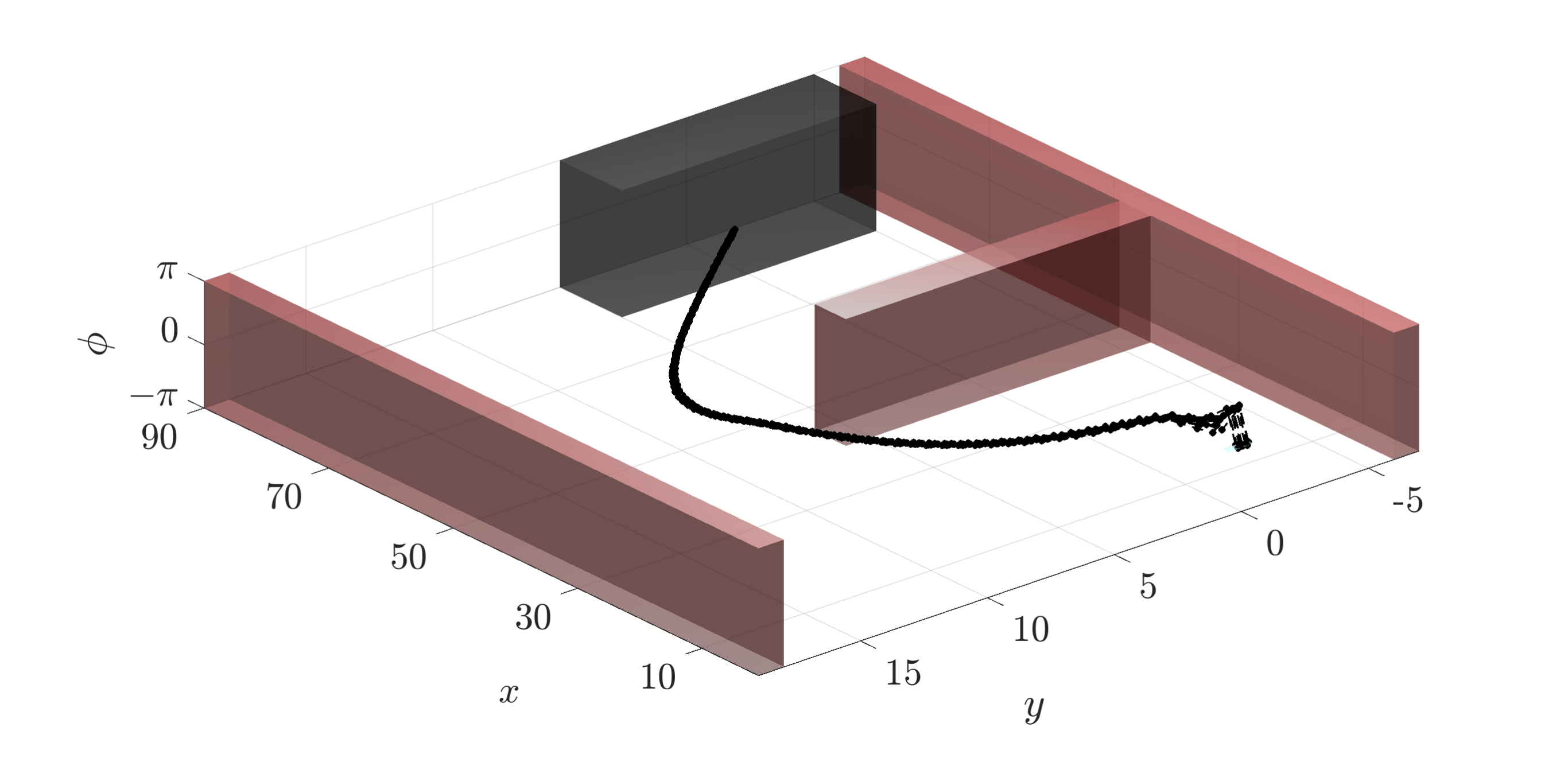}}%
    \put(0.76259538,0.16507632){\color[rgb]{0,0,1}\transparent{0.90118003}\makebox(0,0)[t]{\lineheight{1.25}\smash{\begin{tabular}[t]{c}$\mathbb{X}_0$\end{tabular}}}}%
    \put(0,0){\includegraphics[width=\unitlength,page=2]{CSOvertaking_sim10_3d.pdf}}%
  \end{picture}%
\endgroup%

    \caption{Control performance of the NN controller for the overtaking scenario (Section~\ref{sec:benchmark_overtaking}), demonstrated on 10 random trajectories initialized in $\X_0$ (in blue), trained to avoid the unsafe regions $\X_u$ (in red) and reach a target set (in black).}
    \label{fig:overtaking_nn_controller_sims}
\end{figure}

Now, we move to a system controlled by a NN controller. 
{For this, consider the overtaking scenario in Example~\ref{exmp:overtaking}, visualized in Figures~\ref{fig:safety_setup} and \ref{fig:overtaking_nn_controller_sims}.}
The NN controller {$\pi\colon\X\rightarrow\U$ was trained using a \emph{twin-delayed deep deterministic} (TD3) policy gradient RL algorithm and} features two layers of 200 neurons each and issues a steering wheel angle $u_t\in[-\pi,\pi]$.
We set a fixed velocity $v=1$.
The goal is to certify that the controlled system remains safe by computing the probability that the system initialized in $[{x}_0,{y}_0,\phi_0]\T\in \X_0$ does not enter the unsafe region $\X_u$ within {$T=5$} time steps.

\paragraph{{Setup.}}
The LP is generated based on a lattice {${\smash{\Theta_{\tilde{N}}}}$ of size $\tilde{N}= 70^3$.} We select a spectral basis of $M={5}^3-1$ wavenumbers, resulting in a barrier characterized by $2M+1={249}$ coefficients $b_0,b_1,\ldots\in\R$. The kernel hyperparameters are selected to be {$\sigma_f=7$ and $\sigma_l=[0.054, 0.094, 5.078]$ for the input kernel $k_x$ and $\sigma_l=[0.525, 0.05, 0.525]$ for the output kernel $k_+$}.
{For the computation of the constraint-tightening coefficients, we inflate the sets by 3\% (cf. Remark~\ref{rem:set_scaling}). Throughout all experiments, we set $\bar\B=6.1$.}

\begin{figure}
     \centering
     \begin{subfigure}{0.49\columnwidth}
         \centering
         \includegraphics[width=\textwidth]{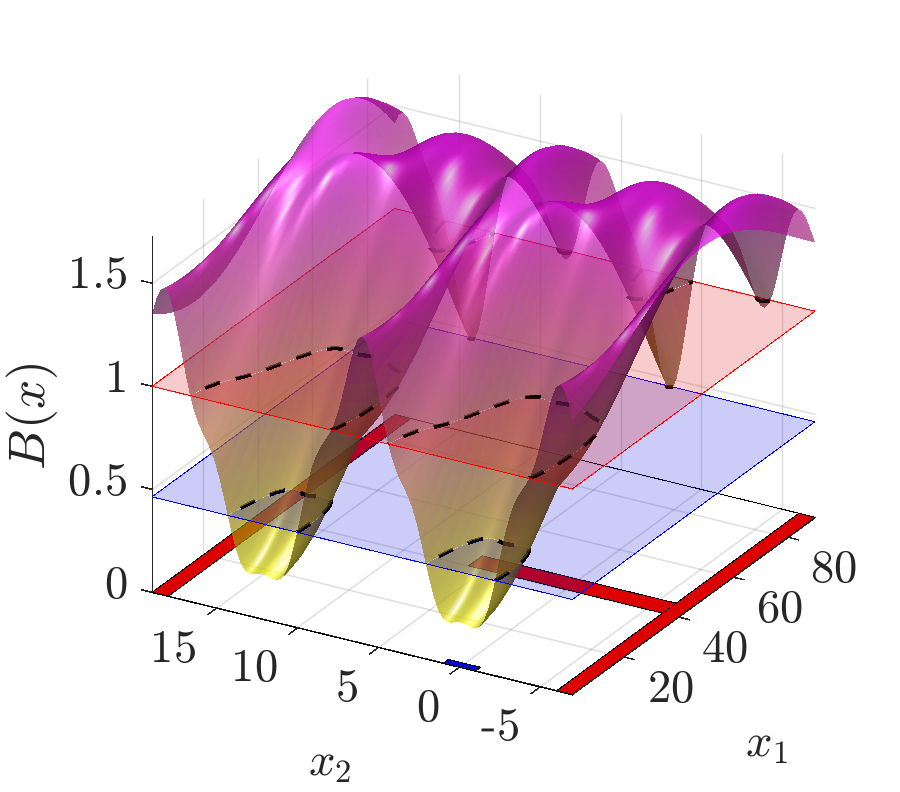}
         \caption{Empirical CBC ($\varepsilon\bar{\B}\sigma_f=0$), $p^\pi_N={53.7}\%$}
         \label{fig:barrier_csovertaking_empirical}
     \end{subfigure}
     \hfill
     \begin{subfigure}{0.49\columnwidth}
         \centering
         \includegraphics[width=\textwidth]{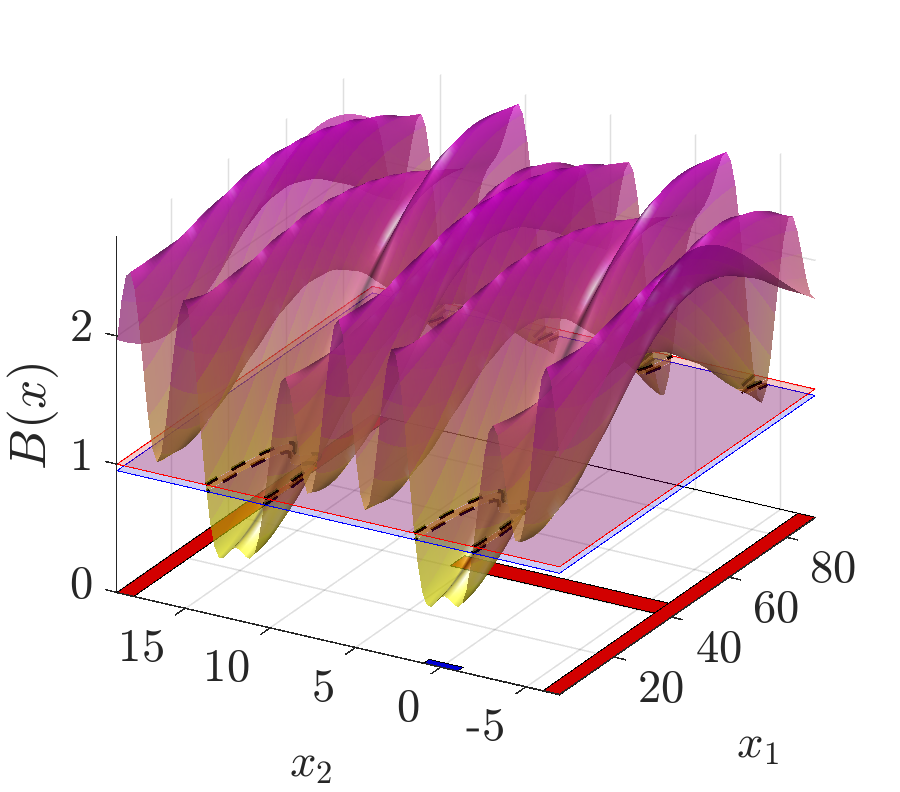}
         \caption{Robust CBC {($\varepsilon\bar{\B}\sigma_f=0.897$), $p^\pi_N=5.07\%$}}
         \label{fig:barrier_csovertaking_robustified}
     \end{subfigure}
        \caption{CBCs and level sets {at} ${\B(x)=1}$ (in red) and $\eta$ (in blue) computed for the overtaking scenario (Section~\ref{sec:benchmark_overtaking}) shown for a fixed $\phi=0$,
        (a) based only on the empirical data, and
        (b) robustified with {$\varepsilon=0.021$ and $\bar{\B}=6.1$.}
        We indicate the intersections of the barrier and the level sets as dashed lines.}
\end{figure}

\begin{figure}
    \centering
    \includegraphics[width=\linewidth]{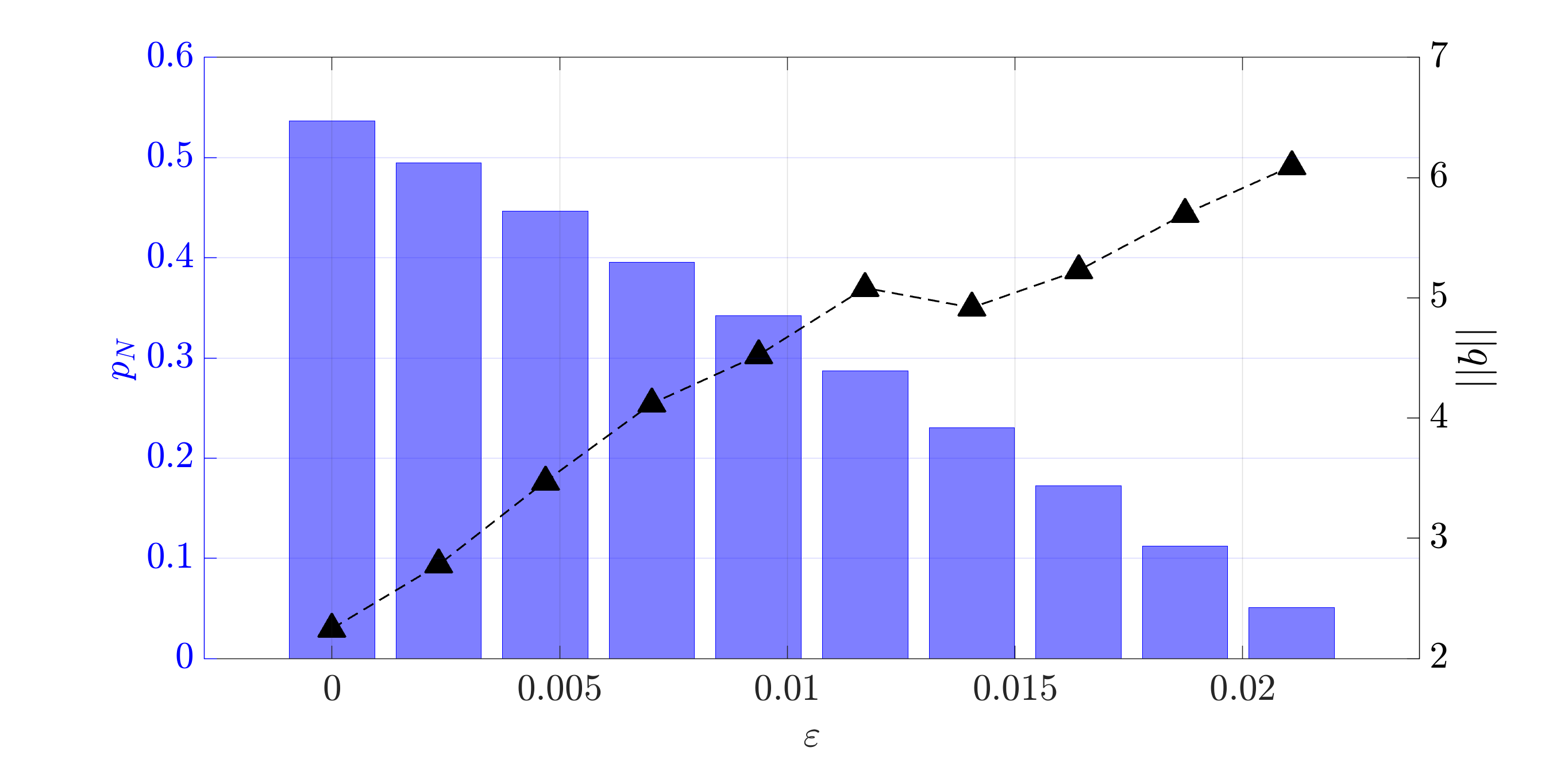}
    \caption{Robust safety probability $p^\pi_N$ (in blue) and required barrier complexity $\norm{b}$ (in black) for the overtaking scenario (Section~\ref{sec:benchmark_overtaking}) generated for different robustness radii $\varepsilon$.}
    \label{fig:probOverRadius2}
\end{figure}

\paragraph{{Results.}}
For $\varepsilon=0$, {$c\approx0$ (within \textsc{Gurobi}'s tolerance)}, and $\eta={0.463}$, we obtain the barrier shown in Figure~\ref{fig:barrier_csovertaking_empirical}, which certifies safety with a probability of at least {$53.7\%$}. 
As {$c\approx0$}, this safety guarantee {holds for time horizons} $T\rightarrow\infty$ (see Remark~\ref{rem:ininite_horizon_safety}).
Note that this is only certifying safety w.r.t. the empirical distribution, i.e., the CME constructed from the observed data.
In order to robustify the result to out-of-distribution behavior, we increase the robustness coefficient $\varepsilon$ and report its influence on the lower bound on the safety probability, $p^\pi_N$, in {Figure~\ref{fig:probOverRadius2}}. Intuitively, increasing $\varepsilon$ tightens the Kushner constraint \eqref{eq:semiinf_prog_kushner} by enforcing the barrier to decay by at least $\varepsilon\bar{\B}\sigma_f$ in every time step.
Note that this is demanding the barrier to be more complex, as observed in an exponentially increasing complexity of the barrier, measured through $\norm{b}$.
Simultaneously, the distance between the level sets ${\B(x)=1}$ and $\eta$ is shrinking, prompting the robust safety probability $p^\pi_N$ to decline.
For the highest reported $\varepsilon={0.021}$, the required decay reaches $\varepsilon\bar{\B}\sigma_f={0.897}$, satisfied by the barrier shown in Figure~\ref{fig:barrier_csovertaking_robustified}, for which the robust safety probability is {$5.07\%$}.

\section{Concluding Remarks}\label{sec:conclusion}

This work contributes to the verification and synthesis of stochastic systems without explicit model knowledge by leveraging the theory of conditional mean embedding and introducing a data-driven approach to control barrier certificates (CBCs). Through the reformulation of probabilistic CBC constraints into a distributionally robust optimization problem, we showed how to obtain the CBC characterization using data with probabilistic correctness guarantees. We showed how the approach can be generalized to temporal logic specifications beyond safety using an automata representation of the specification and the concept of Streett supermartingales. For the data-driven computation of safety barriers, we introduced a finite Fourier expansion to cast the optimization as a linear program.

{
Whilst this paper provides an algorithmic solution for safety \emph{verification}, the resulting linear formulation does not generally extend to black-box control synthesis. In the synthesis setting, the control input enters the constraint nonlinearly through the input kernel, preventing the problem from being recast as a linear program as done here for verification. Only in special cases --- such as systems linear in the control input \cite{santoyo2021barrier} --- a more tractable structure arises. 
In the general case, the synthesis problem may need to be addressed using gradient-based solvers or custom nonlinear global optimizers with deterministic convergence guarantees. Exploring this extension remains an important avenue for future work.
}

Further investigation is needed to reduce the conservatism of the approach and determine sharp ambiguity set radii $\varepsilon$ {with high confidence values $1-\rho$}, including the use of alternative concentration theorems tailored to the class of barrier functions (e.g., via Rademacher complexity) and alternative algorithmic solutions to the Fourier barrier program \eqref{eq:semiinfLP_prog}.
Albeit the presented spectral barrier design lifts the need for an exponential number of spatial support vectors, the number of samples required by the trigonometric sampling bounds remains exponential in the system dimensionality{; the FFT only mitigates this effect. 
It will therefore be interesting to combine the Fourier barrier formulation with nonlinear global optimizers with deterministic convergence guarantees, potentially requiring the development of a custom semi-infinite solver to exploit the problem structure fully. Moreover, techniques such as kernel herding \cite<see, e.g.,>{chen2010kernelherding} could be explored to alleviate the cubic complexity of the CME, although this lies outside the scope of the present work.} Additionally, further exploration of hyperparameter tuning and an optimal selection of the spectral basis are needed.

\acks{
The work of Sadegh Soudjani is supported by the EIC SymAware project 101070802 and the ERC Auto-CyPheR project 101089047. Zhengang Zhong is grateful for the support of the Leverhulme Trust through the Project Award ``Robust Learning: Uncertainty Quantification, Sensitivity and Stability'' (grant agreement RPG-2024-051).

The authors wish to thank Dominik Bongartz for his feedback on maximizing the potential of the MAiNGO solver. We also thank Arthur Gretton for helpful discussions and directing us to valuable references on concentration bounds for CMEs.
We thank Ernesto Casablanca for his efforts in the development of the \lucid toolbox.
}

\appendix

\section{Proof of Lemma~\ref{lem:trig_bound_local}}\label{app:proof_lem_trig_bound_local}
\begin{proof}
    The proof follows the same steps as \citeA[Corollaries~1~\&~2]{pfister2018bounding}. We first prove the upper bound and then the lower bound.

    \medskip
    \noindent\emph{Upper bound:}
    We start by defining the trigonometric polynomial $F\colon\tilde{\X}\to\R$,
    $$F(x):=\B(x)-\frac{\Barrmax_{\tilde{N}}^\S + \Barrmin_{\tilde{N}}^\S}{2},$$
    which is the barrier $\B$ with the output on $\S$ re-centered around zero. 
    $F$ satisfies 
    \begin{equation}
        \max_{x\in\Theta_{\tilde{N}}\cap\S} F(x) = \Barrmax_{\tilde{N}}^\S - \frac{\Barrmax_{\tilde{N}}^\S + \Barrmin_{\tilde{N}}^\S}{2} = \frac{\Barrmax_{\tilde{N}}^\S - \Barrmin_{\tilde{N}}^\S}{2}.\label{eq:F_max}
    \end{equation}
    Via \eqref{eq:separation} and \eqref{eq:F_max}, we have for all $x\in\S$ that
    \begin{align*}
        \B(x) &= F(x) + \frac{\Barrmax_{\tilde{N}}^\S + \Barrmin_{\tilde{N}}^\S}{2},\\
        &\leq \max_{x\in\S} F(x) + \frac{\Barrmax_{\tilde{N}}^\S + \Barrmin_{\tilde{N}}^\S}{2},\\
        &\stackrel{\eqref{eq:separation}}{\leq} 
        C_{\tilde{N}} \max_{x\in\Theta_{\tilde{N}}\cap\S} F(x)
        + A^{\tilde{\X}\setminus\S}_{\tilde{N}}\left(\max_{x\in\Theta_{\tilde{N}}\setminus\S} F(x)-\max_{x\in\Theta_{\tilde{N}}\cap\S} F(x)\right)
        + \frac{\Barrmax_{\tilde{N}}^\S + \Barrmin_{\tilde{N}}^\S}{2},\\
        &\stackrel{\eqref{eq:F_max}}{=} 
        C_{\tilde{N}} \frac{\Barrmax_{\tilde{N}}^\S - \Barrmin_{\tilde{N}}^\S}{2}
        + A^{\tilde{\X}\setminus\S}_{\tilde{N}}\left(\Barrmax_{\tilde{N}}^{\tilde{\X}\setminus\S}-\frac{\Barrmax_{\tilde{N}}^{\S} + \Barrmin_{\tilde{N}}^{\S}}{2}-\frac{\Barrmax_{\tilde{N}}^\S - \Barrmin_{\tilde{N}}^\S}{2}\right)
        + \frac{\Barrmax_{\tilde{N}}^\S + \Barrmin_{\tilde{N}}^\S}{2},\\
        &=
        C_{\tilde{N}} \frac{\Barrmax_{\tilde{N}}^\S - \Barrmin_{\tilde{N}}^\S}{2}
        + A^{\tilde{\X}\setminus\S}_{\tilde{N}}\left(\Barrmax_{\tilde{N}}^{\tilde{\X}\setminus\S}-\Barrmax_{\tilde{N}}^{\S}\right)
        + \frac{\Barrmax_{\tilde{N}}^\S + \Barrmin_{\tilde{N}}^\S}{2},\\
        &=
        \frac{1}{2}\left(\Barrmax_{\tilde{N}}^\S + \Barrmin_{\tilde{N}}^\S + C_{\tilde{N}}\left(\Barrmax_{\tilde{N}}^\S - \Barrmin_{\tilde{N}}^\S\right)\right) + A^{\tilde{\X}\setminus\S}_{\tilde{N}}\left(\Barrmax_{\tilde{N}}^{\tilde{\X}\setminus\S}-\Barrmax_{\tilde{N}}^{\S}\right),
    \end{align*}
    matching the upper bound in Lemma~\ref{lem:trig_bound_local}.

    \medskip
    \noindent\emph{Lower bound:}
    As for the upper bound, we start by defining a trigonometric polynomial,
    \begin{equation}
        F'\colon\tilde{\X}\to\R,
        \quad 
        F'(x):=\frac{\Barrmax_{\tilde{N}}^\S + \Barrmin_{\tilde{N}}^\S}{2} - \B(x), \label{eq:F_2}
    \end{equation}
    which, compared to $F$, is flipped in sign. 
    $F'$ satisfies 
    \begin{equation}
        \max_{x\in\Theta_{\tilde{N}}\cap\S} F'(x) \leq \abs{ \frac{\Barrmax_{\tilde{N}}^\S + \Barrmin_{\tilde{N}}^\S}{2} - \Barrmin_{\tilde{N}}^\S} = \frac{\Barrmax_{\tilde{N}}^\S - \Barrmin_{\tilde{N}}^\S}{2}.\label{eq:F_max_2}
    \end{equation}
    Starting from \eqref{eq:F_2}, we have for all $x\in\S$ that
    \begin{align*}
        \frac{\Barrmax_{\tilde{N}}^\S + \Barrmin_{\tilde{N}}^\S}{2} &= \B(x) + F'(x),\\
        &\stackrel{\eqref{eq:separation}}{\leq} \B(x) + C_{\tilde{N}} \max_{x\in\Theta_{\tilde{N}}\cap\S} F'(x)
        + A^{\tilde{\X}\setminus\S}_{\tilde{N}}\left(\max_{x\in\Theta_{\tilde{N}}\setminus\S} F'(x)-\max_{x\in\Theta_{\tilde{N}}\cap\S} F'(x)\right),\\
        &\stackrel{\eqref{eq:F_max_2}}{=} \B(x) + C_{\tilde{N}} \frac{\Barrmax_{\tilde{N}}^\S - \Barrmin_{\tilde{N}}^\S}{2}
        + A^{\tilde{\X}\setminus\S}_{\tilde{N}}\left(\frac{\Barrmax_{\tilde{N}}^\S + \Barrmin_{\tilde{N}}^\S}{2} - \Barrmin_{\tilde{N}}^{\tilde{\X}\setminus\S} - \frac{\Barrmax_{\tilde{N}}^\S - \Barrmin_{\tilde{N}}^\S}{2}\right).
    \end{align*}
    Reordering yields the upper bound in Lemma~\ref{lem:trig_bound_local}, concluding the proof.
\end{proof}

\section{SILP to LP Relaxation}\label{app:silp_to_lp_derivation}
We provide further details on the finite-constraint relaxation of the SILP \eqref{eq:semiinfLP_prog} to the LP \eqref{eq:linear_prog}.
Let $\{x_0^{(1)},\ldots,x_0^{(\hat{N}_0)}\}\subset\X_0$, $\{x_u^{(1)},\ldots,x_u^{(\hat{N}_u)}\}\subset\X_u$, ${\Theta_{\hat{N}}\allowbreak:=\{x^{(1)},\allowbreak\ldots,\allowbreak x^{(\hat{N})}\}\allowbreak\subset\X}$, and ${\Theta}_{{\tilde{N}}}\allowbreak:=\{x^{(1)},\allowbreak\ldots,\allowbreak x^{({\tilde{N}})}\}\allowbreak\subset{\tilde{\X}}$ be sampling lattices with a common $C_{{\tilde{N}}}$ as in Lemma~\ref{lem:trig_bound}.
Based on the constraint-tightening reasoning outlined in Subsection~\ref{sec:finite_constraint_relaxation}, we have that the first constraint \eqref{eq:semiinfLP_prog_initial} given by $\forall x_0\in\X_0\colon\phi_M(x_0)\T b\leq\eta$ holds if 
\begin{equation*}
    \Barrmax_{{\tilde{N}}}^{\X_0} {\leq} 
    {\frac{
    2\eta + (C_{\tilde{N}}-1)\Barrmin_{{\tilde{N}}}^{\X_0} - 2A^{\tilde{\X}\setminus\X_0}_{\tilde{N}}\Barrmax_{\tilde{N}}^{\tilde{\X}\setminus\X_0}
    }{
    C_{{\tilde{N}}}-2A^{\tilde{\X}\setminus\X_0}_{\tilde{N}}+1
    }}  
\end{equation*}
with the additional variables $\Barrmin_{{\tilde{N}}}^{\X_0}\leq\phi_M(x_0^{(i)})\T b$, $i=1,\ldots,\hat{N}_0$, $\Barrmax_{{\tilde{N}}}^{\X_0}\geq\phi_M(x_0^{(i)})\T b$, $i=1,\ldots,\hat{N}_0$, {and $\Barrmax_{\tilde{N}}^{\tilde{\X}\setminus\X_0}\geq\phi_M(x^{(j)})\T b$, $j=1,\ldots,\tilde{N}$.} 
We call $\Barrmax_{{\tilde{N}}}^{\X_0}=:\hat{\eta}$ and thus get
{\begin{align*}
    \left.\begin{array}{ll}
	    {\Barrmin_{{\tilde{N}}}^{\X_0}\leq}\phi_M(x_0^{(i)})\T b\leq\hat{\eta}, & i=1,\ldots,\hat{N}_0,\\
        {\phi_M(x^{(i)})\T b\leq\Barrmax_{\tilde{N}}^{\tilde{\X}\setminus\X_0},} & {i=1,\ldots,\tilde{N}-\hat{N}_0,}\\
        \textstyle\hat{\eta} := {\frac{ 2\eta + (C_{\tilde{N}}-1)\Barrmin_{{\tilde{N}}}^{\X_0} - 2A^{\tilde{\X}\setminus\X_0}_{\tilde{N}}\Barrmax_{\tilde{N}}^{\tilde{\X}\setminus\X_0}
    }{
    C_{{\tilde{N}}}-2A^{\tilde{\X}\setminus\X_0}_{\tilde{N}}+1
    }} &
    \end{array} \right\}
    \Longrightarrow
    \forall x_0\in\X_0\colon\phi_M(x_0)\T b\leq\eta.
\end{align*}}
Similarly, we obtain for the constraint \eqref{eq:semiinfLP_prog_kushner} that 
{\begin{align*}
    \left.\begin{array}{ll}
	    {\Barrmin_\Delta^{{\X}}\leq}\phi_M(x^{(i)})\T (Hb-b)\leq\hat{\Delta}, &i=1,\ldots,\hat{N},\\
        {\phi_M(x^{(i)})\T (Hb - b)\leq\Barrmax_\Delta^{\tilde{\X}\setminus\X},} & {i=1,\ldots,{\tilde{N}}-\hat{N},}\\
        \hat{\Delta} := {\tfrac{2(c - \varepsilon\bar{\B}\kappa) + (C_{{\tilde{N}}}-1)\Barrmin_\Delta^\X-2A^{\tilde{\X}\setminus\X}_{\tilde{N}}\Barrmax_{\Delta}^{\tilde{\X}\setminus\X}}{C_{{\tilde{N}}}{-2A^{\tilde{\X}\setminus\X}_{\tilde{N}}}+1}} &
    \end{array} \right\}
    \Longrightarrow
    \!\!\!
    \begin{array}{l}
         \forall x\in\X\colon\phi_M(x)\T (Hb-b)  \\
         \hspace{75pt} \leq c - \varepsilon\bar{\B}\kappa. 
    \end{array}
\end{align*}}
For the remaining lower bounds in the constraints \eqref{eq:semiinfLP_prog_unsafe} and \eqref{eq:semiinfLP_prog_initial}, the constraint-tightening reasoning outlined in Subsection~\ref{sec:finite_constraint_relaxation} gives
{\begin{align*}
    &\left.\begin{array}{ll}
	    \hat{\gamma}\leq\phi_M(x_u^{(i)})\T b{\leq\Barrmax_{{\tilde{N}}}^{\X_u}}, & i=1,\ldots,\hat{N}_u,\\
        {\Barrmin_{\tilde{N}}^{\tilde{\X}\setminus\X_u}\leq\phi_M(x^{(i)})\T b,} & {i=1,\ldots,\tilde{N}-\hat{N}_u,}\\
        \textstyle\hat{\gamma} := {\tfrac{2 + (C_{{\tilde{N}}}-1)\Barrmax_{{\tilde{N}}}^{\X_u}{-2A^{\tilde{\X}\setminus\X_u}_{\tilde{N}} \Barrmin_{\tilde{N}}^{\tilde{\X}\setminus\X_u}}}{C_{{\tilde{N}}}{-2A^{\tilde{\X}\setminus\X_u}_{\tilde{N}}}+1}}&
    \end{array} \right\}
    \Longrightarrow
    \forall x_u\in\X_u\colon\phi_M(x_u)\T b\geq{1},\\[.5em]
    &\left.\begin{array}{ll}
	      \hat{\xi}\leq\phi_M(x^{(i)})\T b{\leq\Barrmax_{{\tilde{N}}}^{{{\X}}}}, & i=1,\ldots,\hat{N},  \\
        {\Barrmin_{{\tilde{N}}}^{{\tilde{\X}\setminus\X}}\leq\phi_M(x^{(i)})\T b,} & {i=1,\ldots,{\tilde{N}}-\hat{N},}   \\
        \hat{\xi} := {\tfrac{(C_{{\tilde{N}}}-1)\Barrmax_{{\tilde{N}}}^{{{\X}}}{-2A^{\tilde{\X}\setminus\X}_{\tilde{N}}} \Barrmin_{\tilde{N}}^{\tilde{\X}\setminus\X}}{C_{{\tilde{N}}}{-2A^{\tilde{\X}\setminus\X}_{\tilde{N}}}+1} }&
    \end{array} \right\}
    \Longrightarrow
    \forall x\in\X\colon\phi_M(x)\T b\geq0,
\end{align*}}
respectively.
Substituting the relaxed constraints into the SILP \eqref{eq:semiinfLP_prog}, we obtain the LP \eqref{eq:linear_prog}.

\vskip 0.2in
\bibliography{references}
\bibliographystyle{theapa}

\end{document}